\documentclass[10pt]{article} %

\usepackage{hyperref}
\usepackage{url}

\usepackage[utf8]{inputenc}

\usepackage{amsfonts}
\usepackage{amssymb}
\usepackage{amsmath}
\usepackage{amsthm}
\usepackage{enumerate}
\usepackage{algorithm}
\usepackage{algorithmic}
\usepackage{tikz}
\usepackage{booktabs}
\usepackage{nicefrac}       %
\usepackage{makecell}
\usepackage{array, multirow, boldline}

\newcommand{\E}{\mathbb{E}}

\newcommand{\cW}{\mathcal{W}}

\newcommand{\cN}{\mathcal{N}}

\newcommand{\I}{\mathbb{I}}

\newcommand{\R}{\mathbb{R}}
\newcommand{\norm}[1]{\lvert\lvert #1\rvert\rvert}
\newcommand{\cO}{\mathcal{O}}

\newcommand{\abs}[1]{\lvert #1 \rvert}
\newcommand{\ip}[2]{\langle #1,#2\rangle}

\DeclareMathOperator*{\argmax}{arg\,max}
\DeclareMathOperator*{\argmin}{arg\,min}

\DeclareMathOperator{\erf}{erf}

\newtheorem{theorem}{Theorem}
\newtheorem{assumption}{Assumption}
\newtheorem{lemma}{Lemma}
\newtheorem{definition}{Definition}
\newtheorem{corollary}{Corollary}
\newtheorem{remark}{Remark}

\usepackage{hyperref}
\usepackage{cleveref}

\usepackage{xcolor}

\usepackage{graphicx} %
\usepackage{paralist}
\usepackage{enumitem}

\usepackage[utf8]{inputenc} %
\usepackage[T1]{fontenc}    %
\usepackage{url}            %
\usepackage{booktabs}       %
\usepackage{amsfonts}       %
\usepackage{nicefrac}       %
\usepackage{microtype}      %
\usepackage{xcolor}         %
\usepackage{caption}
\usepackage{subcaption}

\usepackage[accepted]{tmlr}
\let\cite\citep

\title{Collaborative  Compressors  in Distributed Mean Estimation with Limited Communication Budget}
\author{%
\name Harsh Vardhan \email \text{hharshvardhan@ucsd.edu}\\
\addr Department of Computer Science and Engineering\\
University of California, San Diego\\
\authorAND
\name Arya Mazumdar\email \text{arya@ucsd.edu}\\
\addr Halicio{\u{g}}lu Data Science Institute\\
University of California, San Diego\\
}

\def\month{10}  %
\def\year{2025} %
\def\openreview{\url{https://openreview.net/forum?id=AtCKHCoMA7}} %

\begin{document}

\maketitle

\begin{abstract}
    Distributed high dimensional mean estimation is a common aggregation routine used often in distributed optimization methods. Most of these applications call for a communication-constrained setting where vectors, whose mean is to be estimated, have to be compressed before sharing. One could independently encode and decode these to achieve compression, but that overlooks the fact that these vectors are often close to each other. 
    To exploit these similarities, recently Suresh et al., 2022, Jhunjhunwala et al., 2021, Jiang et al, 2023, proposed multiple {\em correlation-aware compression schemes.} However, in most cases, the correlations have to be known for these schemes to work. Moreover,
    a theoretical analysis of graceful degradation of these correlation-aware compression schemes with increasing {\em dissimilarity} is limited to only the $\ell_2$-error in  the literature. 
    In this paper, we
    propose four different collaborative compression schemes  that agnostically exploit the similarities among vectors in a distributed setting. 
    Our schemes are all simple to implement and computationally efficient, while resulting in big savings in communication. The analysis of our proposed schemes show how the $\ell_2$, $\ell_\infty$ and cosine estimation error varies with the degree of similarity among vectors. %
\end{abstract}

\section{Introduction}

\begin{figure}[t!]
    \centering
    \begin{subfigure}[t]{0.49\textwidth}
    \centering
        \includegraphics[width=\textwidth]{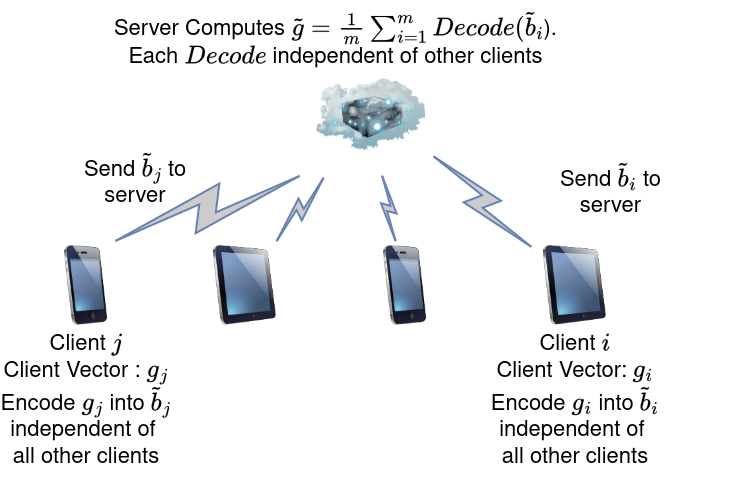}
        \caption{Independent Compression}
        \label{fig:indep_compress}
    \end{subfigure}
    \hfill
    \begin{subfigure}[t]{0.49\textwidth}
        \includegraphics[width=\textwidth]{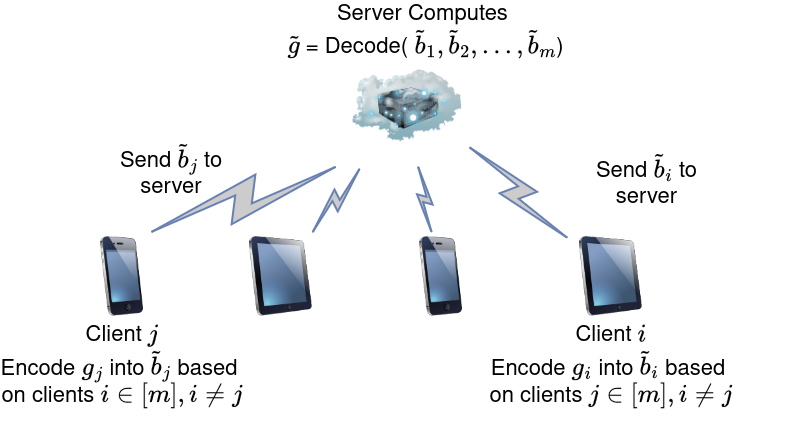}
        \caption{Collaborative Compression}
        \label{fig:collab_compress}
    \end{subfigure}
    \caption{Compression for Distributed Mean Estimation}
    \label{fig:compress_dme}
\end{figure}

We study the problem of estimating the empirical mean, or average, of a set of high-dimensional vectors in a communication-constrained setup.
We assume a distributed problem setting, where $m$ clients, each with a vector $g_i \in \R^d$, are connected to a single server (see, Fig.~\ref{fig:indep_compress}). Our goal is to estimate their mean $g$ on the server, where
\begin{align}
    g \triangleq \frac{1}{m}\sum_{i\in [m]} g_i .
\end{align}
We use $[m]$ to denote the set $ \{1,2,\ldots,m\}$.
The clients can communicate with the server via a communication channel which allows limited communication. 
The server does not have access to data but has relatively more computational power than individual clients. 

This problem, referred to as {\em distributed mean estimation} (DME), is an important subroutine in several distributed learning applications. One common application is distributed training
or federated learning~\cite{mcmahan2016communication,mcmahan2017federated}, when different clients correspond to different edge devices. %

The typical learning task for DME is supervised learning via gradient-based methods~\cite{bottou_tradeoffs_2007,sgd}. In this case, the vectors $g_i$ correspond to the gradient updates for each client $i$ computed on its local training data, and $g$ is the average gradient over all clients. On the other hand, distributed mean estimation is also used in unsupervised learning problems such as distributed KMeans~\cite{Liang2013DistributedPA}, distributed PCA~\cite{liang2014improved} or distributed power iteration~\cite{li2021communication}. In distributed KMeans and distributed power iteration, $g_i$ corresponds to estimates of the cluster center and the top eigenvector, respectively, on the $i^{th}$ client.

The naive strategy %
 of clients sending their vectors $g_i$ to the server for DME incurs no error; however, it has a high communication cost, rendering it burdensome in most of the real-world network applications. A principled way to tackle this is to use compression: each client $i\in [m]$ compresses its vector $g_i$ into an efficient encoding $\tilde{b}_i \in \mathcal{B}_i$ which can then be sent to the server; The server forms an estimate $\tilde{g}$ of the mean $g$ using the encodings $\{\tilde{b}_i\}_{i\in [m]}$. We can then compute the error of the estimate $\tilde{g}$ and the number of bits required to communicate $\tilde{b}_i$ (i.e., $\log_2 |\mathcal{B}_i|$) to analyze the efficiency of the compression scheme. As opposed to distributed statistical inference~\cite{stat_est_1, garg_communication_2014}, we do not assume that $g_i$ are sampled from a distribution, and instead the estimation error of these schemes is computed in terms of $g_i$.

One way to approach this compression paradigm is when each client compresses its vector oblivious to others, and the server separately decodes the vectors before aggregating (Figure~\ref{fig:indep_compress}). We call this {\em independent compression} and several existing works~\cite{konecny_randomized_2018,suresh_distributed_2017,safaryan_kashin, vqsgd, vargaftik_drive_2021,ghosh2021communication} use such a compression scheme. %
The simplest example of this scheme is RandK~\cite{konecny_randomized_2018}, where each client sends only $K \in \mathbb{N}$ coordinates as $\tilde{b}_i$, and the server estimates $\tilde{g}$ as the average of $K$-sparse vectors from each client. As $K < d$, this scheme requires less communication than sending the full vector $g_i$ from each client $i\in [m]$. %

However, independent compressors suffer from a significant drawback, especially when the vectors to be aggregated are similar/not-too-far, which is often the case for gradient aggregation in distributed learning. Consider the case when two distinct clients $i,j\in [m]$ have different vectors $g_i \neq g_j$, but they differ in only one coordinate. Then, independent compressors like RandK will end up sending $\tilde{b}_i$ and $\tilde{b}_j$ which are very similar (in fact, same with high probability) to each other, and therefore wasting communication. %
Collaborative compressors~\cite{suresh_correlated_2022,szlendak_permutation_2021,jhunjhunwala_leveraging_2021,jiang_correlation_2023} can alleviate this problem. Figure~\ref{fig:collab_compress} describes a collaborative compressor, where the encodings $\{\tilde{g}_i\}_{i\in [m]}$ may not be independent of each other and a decoding function {\em jointly} decodes all encodings to obtain the mean estimate $\tilde{g}$. Clearly, this opens up more possibilities 
to reduce communication - but also the error of collaborative compressors can be made to scale as the variability  of the vectors. %

The amount of required communication also depends on the metric for estimation error. Among the existing schemes for collaborative compressors, most provide guarantees on  the $\ell_2$ error $\norm{\tilde{g} - g}_2^2$~\cite{suresh_correlated_2022,szlendak_permutation_2021,jhunjhunwala_leveraging_2021,jiang_correlation_2023}. Also, in collaborative compressors, the error must ideally be dependent on {\em some measure of correlation/distance} among the vectors, which is indeed the case for all of these schemes. In this paper, the measure of such a distance is denoted with $\Delta$, with some subscript signifying the exact measure; the vectors in question have high similarity as $\Delta \to 0$. The estimation error naturally grows with the dimension $d$, and decays with the number of clients $m$ (due to an averaging).  One of our major contributions is to design a compression scheme that has significantly improved dependence on the number of clients $m$ to counter the effect of growing dimension $d$.

If one were to estimate the unit vector in the direction of the average vector $\frac{1}{m}\sum_{i=1}^m g_i$, which is often important for gradient descent applications, using an estimate of the mean with low $\ell_2$ error can be highly sub-optimal as the $\ell_2$ error might be large even if all the vectors point in the same direction but have different norms. For this, the cosine distance $\arccos (\frac{\ip{\tilde{g}}{g}}{\|\tilde{g}\|\|g\|})$ is a better measure, which has not been studied in the literature. We also give a compression scheme specifically tailored for this error metric. Another interesting metric is the $\ell_\infty$-error which has also not been studied except for in~\cite{suresh_correlated_2022}. There as well, we give an improved dependence of the estimation error on  $m$.

Further drawback of existing collaborative compressors such as,~\cite{jhunjhunwala_leveraging_2021,jiang_correlation_2023} is that they require the knowledge of correlation between vectors before employing their compression. Without this knowledge, their error guarantees do not hold.

\paragraph{Notation.} %
We use $[a, b]$ to denote $[a, a+1, a+2, \ldots, b]$ for $a,b\in \mathbb{N}$. We use $g^{(j)}, j\in [d]$ to denote the $j^{th}$ coordinate of a vector $g\in \R^d$. For a permutation $\rho$   on $[m]$, $\rho^{(i)}$ denotes mapping of $i\in [m]$ under $\rho$. We use $\Pi_m$ to denote the uniform distribution over all permutations of $[m]$. We use $\exp(a)$ to denote $e^{\mathcal{O}(a)}$ for any $a\in \R$. For any matrix $A\in \R^{m \times n}$, $A_k$ for $k\in [m]$ refers to the $k^{th}$ row of $A$.

{\bf Our contributions.} 
We provide four different collaborative compressors, which are communication-efficient,  give error guarantees for different error metrics ($\ell_2$ error, $\ell_\infty$ error and cosine distance), and exhibit optimal dependence on the number of clients $m$ and the diameter of ambient space $B$. To see the advantage of collaboration, we define few natural similarity metrics. 
All our schemes show graceful degradation of error with the similarity metric between different clients. %
Our schemes have three subroutines: $\texttt{Init}$ which corresponds to initialization/setting up a protocol, $\texttt{Encode}$ which is performed individually at each client to obtain their encoding $\tilde{b}_i$, and $\texttt{Decode}$ which is performed at the server on all the encodings to obtain estimate of mean $\tilde{g}$. 

\begin{table*}[t!]
\small
\centering

    \begin{tabular}{cccc}
    \toprule
    Compressor & Error metric &  Error &  \# Bits/client \\
    \midrule
    \makecell{NoisySign\\ (Algorithm~\ref{alg:noisy_sign})}& $\norm{\tilde{g} - g}_\infty$ & $ \exp(\frac{\norm{g}_\infty^2}{\sigma^2})\cdot(\Delta_{\phi} + \sqrt{\frac{\log m}{m}})$& $d$\\
    \makecell{HadamardMultiDim \\ (Algorithm~\ref{alg:hadamardmultidim})} & $\E[\norm{\tilde{g} - g}_\infty]$ & $\frac{B}{2^{m-1}} + \Delta_{\rm Hadamard}$ & $d$\\
    \makecell{SparseReg \\ (Algorithm~\ref{alg:sparc_1}) } & $\E[\norm{\tilde{g} - g}_2^2]$ & $B^2\exp\left(-\frac{2m\log L}{d}\right) + \Delta_{\rm reg}$ & \makecell{$\log L$\\ ($L\ge 1$ tunable)}\\
    \makecell{OneBit \\ (Algorithm~\ref{alg:onebit})} & $\arccos{\ip{\tilde{g}}{g}}$ & \makecell{%
    $\pi(\Delta_{\rm corr} + \frac{d}{mt})$ %
   } & \makecell{$t$\\ ($t\ge 1$ tunable)}\\
    \bottomrule
    \end{tabular}
    \caption{\small Theoretical results for our proposed collaborative compression schemes. $\Delta_{\Phi}, \Delta_{\rm Hadamard},\Delta_{\rm reg}$ and $\Delta_{\rm corr}$ are measures of average dissimilarity between vectors $\{g_i\}_{i\in [m]}$  defined in Theorems~\ref{lem:sign_error}, \ref{thm:hadamard_error}, \ref{thm:sparsereg} and Lemma~\ref{lem:label flip} respectively. For NoisySign, $\sigma >0$ is an algorithm parameter. For HadamardMultiDim, we assume $\norm{g_i}_\infty \leq B, \forall i \in [m]$. For SparseReg, we assume $\norm{g_i}_2 \leq B, \forall i \in [m]$ and $L$ is an algorithm parameter. For OneBit, $g$ is the unit vector along the average $\frac{1}{m}\sum_{i=1}^m g_i$ and $\tilde{g}$ is also a unit vector.}
    \label{tab:theoretical_results}
\end{table*}

Below our main contributions are summarized. The theoretical guarantees for our algorithms are provided in Table~\ref{tab:theoretical_results}. For each of our proposed compressors, increasing $m$ increases the total communication required (which is a product of $m$ and the $\#$ Bits/client), but reduces the estimation error in the appropriate error metric up to a constant $\Delta$ that depends on the deviation between client vectors.

\begin{enumerate}[wide, labelwidth=!,labelindent=0pt]
        \item We provide a simple collaborative scheme based on the popular signSGD~\cite{bernstein2018signsgd} scheme, NoisySign (Algorithm~\ref{alg:noisy_sign}), where  sign of each coordinate of a vector is sent after adding Gaussian noise. An advantage of this scheme, compared to others is that we can infer the vector $g$ with an $\ell_\infty$ error guarantee increasing with $\norm{g}_\infty$ and decreasing with $m$, without the knowledge of $\norm{g}_\infty$ itself. The dissimilarity  is $\Delta_{\Phi} = \mathcal{O}(\frac{1}{m\sigma}\sum_{i=1}^m \norm{g - g_i}_\infty)$, where $\sigma$ is the variance of the noise added (Theorem~\ref{lem:sign_error}). This scheme is described in \Cref{sec:noisy_sign}.
    \item {\bf ($\ell_\infty$-guarantee)} For vectors with $\ell_\infty$ norm bounded by $B$, we propose a collaborative compression scheme, HadamardMultiDim (Algorithm~\ref{alg:hadamardmultidim}) which performs coordinate-wise collaborative binary search. We obtain the best dependence on $m$ and $B$ for the $\ell_\infty$ error ($\mathcal{O}(B\cdot \exp(-m))$) while suffering from an extra error term $\Delta_{\rm Hadamard}$, which is a measure of average dissimilarity between compressed vectors. $\Delta_{\rm Hadamard}$  lies in the range $[\Delta_{\infty}, \Delta_{\infty, \max}]$ where $\Delta_{\infty} = \max_{j\in [d]}\frac{1}{m}\sum_{i=1}^m \abs{g^{(j)}_i - g^{(j)}}$ and $\Delta_{\infty,\max} = \max_{j\in [d], i\in [m]}\abs{g^{(j)}_i - g^{(j)}}$ (Theorem~\ref{thm:hadamard_error}). In Section~\ref{sec:example}, we provide a practical example where value of $\Delta_{\mathrm{Hadamard}}$ can be approximated and use it compare theoretical guarantees of HadamardMultiDim with those of baselines in Table~\ref{tab:related_works}. 
    \item {\bf ($\ell_2$-guarantee)} For vectors with $\ell_2$ norm bounded by $B$, we provide a collaborative compression scheme SparseReg (Algorithm~\ref{alg:sparc_1}) based on Sparse Regression Codes~\cite{venkataramanan_lossy_2014,venkataramanan_lossy_2014-1}. We obtain the best dependence on $B$ and $m$ for the $\ell_2$ error ($\mathcal{O}(B\exp(-m/d))$)
    while compressing to much less than $d$ bits (in fact, to a constant number of bits) per client. %
    The error consists of a penalty for the dissimilarity, $\Delta_{\rm reg}$, the average dissimilarity between compressed vectors which lies in the range $[\Delta_{2}, \Delta_{2,\max}]$ where $\Delta_2= \frac{1}{m}\sum_{i=1}^m \norm{g - g_i}_2^2$ and $\Delta_{2,\max} = \max_{i\in [m]}\norm{g - g_i}_2^2$ (see, Theorem~\ref{thm:sparsereg}).
    \item {\bf (cosine-guarantee)} For unit norm vectors $\{g_i\}_{i\in [m]}$, we estimate the unit vector $g$ in the direction of the average $\frac{1}{m}\sum_{i=1}^m g_i$. %
    For this, motivated by one-bit compressed sensing~\cite{boufounos20081}, our collaborative compression scheme, OneBit (Algorithm~\ref{alg:onebit}), sends the sign of the inner product between the vector $g_i$ and a random Gaussian vector. By establishing an equivalence to halfspace learning with malicious noise, we propose two decoding schemes: the first one is based on ~\cite{pmlr-v206-shen23a} which is optimal for halfspace learning but harder to implement and a second one, based on ~\cite{kalai_agnostically_2008} which is easy to implement. Both schemes are computationally efficient, and have an extra dissimilarity term in the error,  $\Delta_{\rm corr} = \frac{1}{m\pi}\sum_{i=1}^m \cos^{-1}(\ip{g}{g_i})$, which is the appropriate dissimilarity between unit vectors (see Theorem~\ref{lem:one_bit_optimal_bias}). 
    \item {\bf (Experiments)} We perform a simulation for DME with our schemes as the dissimilarities vary and compare the three different error metrics from above with various existing baselines (Fig~\ref{fig:gaussian}-\ref{fig:unit_norm}).  We also used our DME subroutines in the downstream tasks of KMeans, power iteration, and linear regression on real (and federated) datasets (Fig~\ref{fig:kmeans_mnist}-\ref{fig:linreg_synthetic}). Our schemes have lowest error in all metrics for low dissimilarity regime.
\end{enumerate}
\begin{table*}[t!]

\small
\begin{center}
\resizebox{\columnwidth}{!}{
\begin{tabular}{cccc}
\toprule
 Compressor  &  Error & \# Bits/client &  Notes \\
\midrule
RandK~\cite{konecny_randomized_2018} &  $\mathcal{O}(\frac{d}{K}\tilde{B}^2)$ & $32K  + K\log d$& Independent\\
SRQ~\cite{suresh_distributed_2017}& $\mathcal{O}(\frac{\log d}{m(K-1)^2}\tilde{B}^2)$& $Kd$&Independent\\
Kashin~\cite{safaryan_kashin} &  $\mathcal{O}(\left(\frac{10\sqrt{\lambda}}{\sqrt{\lambda}-1)}\right)^4\tilde{B}^2)$&$31 + \lambda d$ & Independent\\

Drive~\cite{vargaftik_drive_2021} & $\mathcal{O}( \tilde{B}^2)$& $32 + d$ & Independent \\
PermK~\cite{szlendak_permutation_2021}& $\mathcal{O}((1 - \max\{0, \frac{m-d}{m-1}\})\Delta_{2})$ & $32K     + K\log d$ & Collaborative\\
RandKSpatial~\cite{jhunjhunwala_leveraging_2021}&$\mathcal{O}(\frac{d}{mK}\Delta_{2})$ & $32K + K\log d$& Needs Correlation\\
RandKSpatialProj~\cite{jiang_correlation_2023}&$\mathcal{O}(\frac{d}{mK}\Delta_{2})$ & $32K + K\log d$& Needs Correlation\\
Correlated SRQ~\cite{suresh_correlated_2022}& $\mathcal{O}\left(\frac{1}{m}\min\{\frac{\sqrt{d} \Delta_{\infty}^d B}{K},\frac{d B^2}{K^2}\}\right)$ & $2 d \log K +K\log d$& $\norm{g_i}_2 \leq B,\forall i\in [m]$\\
\bottomrule
\end{tabular}
}
\end{center}

\caption{\small Comparison of existing independent and collaborative compressors in terms of $\ell_2$ error and bits communicated. $K$ is the number of coordinates communicated for sparsification methods(RandK, PermK, RandKSpatial, RandKSpatialProj) and the number of quantization levels for quantization methods (SRQ, vqSGD, Correlated SRQ). The constant $\lambda$ is a parameter of the Kashin scheme. Further, $\tilde{B}^2 =  \frac{1}{m}\sum_{i=1}^m \norm{g_i}_2^2, \Delta_{2} = \frac{1}{m}\sum_{i=1}^m \norm{g_i - g}_2^2$, and $\Delta_{\infty} = \max_{j\in [d]}\frac{1}{m}\sum_{i=1}^m \abs{g_i^{(j)} - g^{(j)}}$. It is also assumed that a real is equivalent to $32$ bits, which is an informal norm in this literature. %
}
\label{tab:related_works}
\end{table*}

\paragraph{Organization.} In the next subsection, we present related works in distributed mean estimation. The NoisySign algorithm is given in \Cref{alg:noisy_sign}, and its analysis can be found in  \Cref{sec:noisy_sign}. In Section~\ref{sec:optimal_m}, we present the two schemes obtaining optimal  dependence on $m$, HadamardMultiDim in Subsection~\ref{sec:hadamard} and SparseReg in Subsection~\ref{sec:sparsereg}. In Section~\ref{sec:one_bit}, we analyze the OneBit compression scheme. Finally, in Section~\ref{sec:experiments}, we provide experimental results for our schemes.

\subsection{Related Works}

\label{sec:related_works}
{\bf Compressors in Distributed Learning.} Starting from ~\cite{konevcny2016federated} most compression schemes in distributed learning involve either quantization or sparsification. In quantization schemes, the real valued input space is quantized to specific  levels, and each input is mapped to one of these quantization levels. A theoretical analysis for unbiased quantization was provided in ~\cite{alistarh_qsgd_2017}.  Subsequently,  the distributed mean estimation problem with limited communication was formulated in ~\cite{suresh_distributed_2017} where two schemes, stochastic rotated quantization (SRQ) and variable length coding, were proposed. These schemes matched the lower bound for communication and $\ell_2$ error in terms of $\tilde{B}^2 = \frac{1}{m}\sum_{i=1}^m \norm{g_i}_2^2$. Performing a coordinate-wise sign is also a quantization operation, introduced in ~\cite{pmlr-v80-bernstein18a}. Further advances in quantization include multiple quantization levels~\cite{terngrad}, probabilistic quantization with noise~\cite{chen_distributed_2020, jin_stochastic-sign_2021, safaryan_stochastic_2021}, vector quantization~\cite{vqsgd,zhu2023optimal}, and applying structured rotation before quantization~\cite{vargaftik_drive_2021,safaryan_kashin}. Sparsification involves selecting only a subset of coordinates to communicate. Common examples include RandK~\cite{konecny_randomized_2018}, TopK~\cite{stich_sparsified_2018} and their combinations~\cite{beznosikov_biased_2022}. Note, for all independent compressors, the $\ell_2$ error scales as $\tilde{B}^2$.

{\bf Collaborative Compressors.}
PermK~\cite{szlendak_permutation_2021} was the first collaborative compressor, where each client would send a different set of $K$ coordinates. Their error scales with the empirical variance, $\Delta_2 = \frac{1}{m}\sum_{i=1}^m \norm{g_i - g}_2^2$. If $\Delta_2$ is known, or one of the vectors $g_i$ is known, the lattice-based quantizer in ~\cite{davies_new_2020}  and correlated noise based quantizer in ~\cite{mayekar_wyner-ziv_2021} obtains $\ell_2$ error in terms of $\Delta_2$. Further, RandKSpatial~\cite{jhunjhunwala_leveraging_2021}
 and RandKSpatialProj~\cite{jiang_correlation_2023} utilize the correlation information to obtain the correct normalization coefficients for RandK with rotations, obtaining guarantees in terms of $\Delta_2$. In absence of correlation information, they propose a heuristic. A quantizer also based on correlated noise,  was proposed in~\cite{suresh_correlated_2022} which achieves the lower bound for scalars. However, for $d$-dimensional vectors of $\ell_2$-norm at most $B$, their dependence on dimension $d$ and number of clients $m$ can be improved by our schemes.

We provide a summary of existing compressors in Table~\ref{tab:related_works}, along with their error guarantees.

\section{NoisySign for unbounded $\norm{g_i}_\infty$}
\label{sec:noisy_sign}

In this section, we utilize collaborative compression to use sign compressor when $\norm{g_i}_\infty$ is unbounded. The sign-compressor~\cite{bernstein2018signsgd} applies the ${\rm sign}$ function coordinate-wise, where ${\rm sign}(x) = +1$ if $x\geq 0$ and $-1$ otherwise. For this section, we will focus on a single coordinate $j\in [d]$. Note that for any $i\in [m]$, ${\rm sign}(g_i^{(j)})$ does not have information about $\abs{g_i^{(j)}}$. Existing compressors~\cite{pmlr-v119-karimireddy20a} remedy this by sending $\abs{g_i^{(j)}}$ separately, or assuming that $\abs{g_i^{(j)}}$ is bounded by some constant $B$~\cite{safaryan_stochastic_2021,jin_magnitude_2023,tang_z-signfedavg_2023}. In the second case, the maximum error that can be incurred is $\frac{B}{2}$. This can be improved by adding uniform symmetric noise before taking signs~\cite{chen_distributed_2020,chzhen_signsvrg_2023} depending on $g$. However, if no information is available about $\abs{g_i^{(j)}}$, we cannot provide an estimate of $g_i^{(j)}$.

We utilize the concept of adding noise before taking signs, however, to accommodate possibly unbounded $\abs{g_i^{(j)}}$, we add symmetric noise with unbounded support. One choice for such noise  is the Gaussian distribution $\cN(0,\sigma^2)$ for some $\sigma\in\R$. For $\xi_i^{(j)} \sim \cN(0,\sigma^2)$, we send  $\tilde{b}_i^{(j)} = {\rm sign}(g_i^{(j)} + \xi_i^{(j)})$ as the encoding. Note that $\E[\tilde{b}_i^{(j)}] = \Phi_\sigma(g_i^{(j)})$, where $\Phi_\sigma(t) = 2\Pr_{x\sim \cN(0, \sigma^2)}[x \geq -t] - 1 = {\rm erf}(\frac{t}{\sqrt{2}\sigma})$, and ${\rm erf}$ is the error function for the unit normal distribution.  A single $\tilde{b}_i^{j}$ gives us information about $g_i^{(j)}$, however, using it to decode $g_i^{(j)}$ might incur a very large variance. However, assuming that all $g_i^{(j)}$ are close to $g^{(j)}$ for $i\in [m]$, $\frac{1}{m}\sum_{i=1}^m \tilde{b}_i^{(j)}$ is a good estimator for $\Phi_\sigma(g^{(j)})$. So, to estimate $g^{(j)}$, we can use $\Phi_\sigma^{-1}(\frac{1}{m}\sum_{i=1}^m \tilde{b}_i^{(j)})$. This scheme performed coordinate-wise is the NoisySign algorithm described in Algorithm~\ref{alg:noisy_sign}. We provide estimation error for recovering $g$ using this scheme.

\begin{theorem}[Estimation error of noisy sign]\label{lem:sign_error}
 With probability $1-2dm^{-c}$, for some constant $c>0$, the estimation error of  ~\Cref{alg:noisy_sign} is
\begin{align}
    &\norm{\tilde{g} - g}_\infty \leq \sqrt{\frac{\pi}{2}}\left(\left(1 - \frac{\Delta_\Phi + \sqrt{\frac{2c\log m }{m}}}{\alpha(\norm{g}_\infty)}\right)^{-1} - 1\right),
\end{align}
where  $\Delta_\Phi \triangleq  \max_{j\in [d]}\abs{\frac{1}{m} \sum_{i=1}^m \Phi_\sigma(g_i^{(j)}) - \Phi_\sigma(g^{(j)})}$ and $\alpha(u) \triangleq 1 - \Phi_\sigma(u)$. Further, for small $\Delta_{\phi}$ and $\frac{\norm{g}_\infty}{\sigma}$, and large $m$,  
    \begin{align}\label{corr:noisy_sign}
        \norm{\tilde{g} - g}_{\infty} = \mathcal{O}\left(\exp( \norm{g}_\infty^2/\sigma^2)\cdot(\Delta_{\phi} + \sqrt{\log m/ m})\right).
    \end{align}
\end{theorem}
Applying $\Phi_\sigma^{-1}$ to estimate $g$ makes our scheme collaborative, as an independent scheme would have individually decoded each client's $\tilde{b}_i^j$ and thus would not take advantage of concentration over $m$ machines. To gain insight into the error, note that $(1-x)^{-1} -1 \approx x,$ for small $x$. The error increases with the increase in $\norm{g}_\infty$ as we are compressing unbounded variables $g_i$ into the bounded domain $[-1,1]$ which is the range of the function $\Phi_\sigma$.The number of clients $m$ determines the resolution with which we can measure on this domain, as the value $\frac{1}{m}\sum_{i=1}^m \tilde{b_i}$ can only be in multiples of $\frac{1}{m}$. Therefore, increasing $m$ decreases the error.
   As $m \to \infty,$ the $\ell_\infty$-error approaches $\Delta_\Phi \exp(\frac{\norm{g}_\infty}{\sigma^2})$. Note that $\Delta_\Phi$ determines the average separation between vectors in terms of the $\Phi_\sigma$ operator. If vectors $g_i$ are similar to each other, $\Delta_\Phi$ is small, and the error is small as a result. Further, $\Delta_\Phi$ can be bounded by more interpretable quantities if the average separation between $g_i$ and $g$ is small:
\begin{align}\label{rem:noisy_sign_delta}
    \Delta_\Phi \leq \sqrt{\frac{2}{\pi}}\frac{1}{m\sigma}\sum_{i\in [m]} \norm{g_i - g}_\infty.
\end{align}
Note that $\Delta_\Phi$ is always $\leq 2$, so if the average error in terms of $\ell_\infty$ norm is much smaller than $\sigma$, then the above bound makes sense. Combining Eq~\eqref{rem:noisy_sign_delta} with Eq~\eqref{corr:noisy_sign}, we can find an asymptotic expression for the estimation error. A large value of $\sigma$ can help us reduce this upper bound on $\Delta_\phi$ as well as the impact of $\alpha(\norm{g}_\infty)$ on the estimation error. One can tune the value of $\sigma$ if additional information is known, for instance a large upper bound on $\norm{g}_\infty$ and $\frac{1}{m}\sum_{i=1}^m \norm{g_i - g}_\infty$. However, even without this information, we can run our algorithm. In contrast, the value of $\max_{i\in [m]}\norm{g}_\infty$ is required to even run vanilla sign compression algorithms. Additionally, it yields a  constant error of $\mathcal{O}(\max_{i\in [m]}\norm{g_i}_\infty)$, as each sign would need to be accurate. If an upper bound is known for $\max_{i\in [m]}\norm{g_i}_\infty$, then the error scales with this upper bound. In contrast, for large $m$ and small $\Delta_{\Phi}$, knowledge of a large upper bound helps us to set a large $\sigma$ and thus our collaborative compressor performs much better. Proofs for this section are provided in \Cref{sec:noisy_sign_proofs}.
\begin{minipage}{0.48\textwidth}
\begin{algorithm}[H]
\centering
    \caption{NoisySign}
    \label{alg:noisy_sign}
    \begin{algorithmic}
        \STATE \texttt{\underline{Encode($g_i$)}}
        \STATE Sample $\xi_i \sim \cN(0, \sigma^2\I_d)$
        \STATE $\tilde{b_i} = {\rm sign}(g_i + \xi_i)$
        \RETURN $\tilde{b_i}$.
        \STATE \texttt{\underline{Decode($\{\tilde{b_{i}}\}_{i\in [m]}$)}}        
        \STATE $\tilde{g}^{(j)} \gets \Phi_{\sigma}^{-1}(\frac{1}{m}\sum_{i=1}^m \tilde{b_{i}}^{(j)}), j =1, \dots, d$
        \RETURN $\tilde{g}$
    \end{algorithmic}
\end{algorithm}        
\begin{algorithm}[H]
\caption{Hadamard1DEnc}
\label{alg:hadamard1denc}
\begin{algorithmic}
\STATE {\bfseries Input:} Scalar $s$, Level $K$
\STATE $S_{K}^{-} = \cup_{k=0}^{2^{K-1} -1}[-B + \frac{2kB}{2^{K-1}},-B + \frac{(2k+1)B}{2^{K-1}}]$
\RETURN $-1$ if $s \in S_{K}^{-}$ else $+1$
\end{algorithmic}
\end{algorithm}

\end{minipage}%
\hfill
\begin{minipage}{0.48\textwidth}
\begin{algorithm}[H]
    \caption{HadamardMultiDim}
    \label{alg:hadamardmultidim}
    \begin{algorithmic}
        \STATE \texttt{\underline{Init()}}
        \STATE Sample a random permutation $\rho \sim \Pi_m$.
        \STATE Clients and server share $\rho$.
        \STATE \texttt{\underline{Encode($g_i$)}}
        \FOR{$j\in [d]$}
            \STATE $\tilde{b_i}^{(j)} \gets {\rm Hadamard1DEnc}(g_i^{(j)}, \rho^{(i)})$
        \ENDFOR
        \RETURN $\tilde{b_i}$
        \STATE \texttt{\underline{Decode($\{\tilde{b_i}\}_{i\in [m]}$)}}
        \FOR{$j\in [d]$}
            \STATE $\tilde{g}^{(j)} = \sum_{i=1}^m \tilde{b_i}^{(j)}\cdot\frac{B}{2^{\rho^{(i)} - 1}}$
        \ENDFOR
        \RETURN $\tilde{g}$
    \end{algorithmic}
\end{algorithm}

\end{minipage}%

For NoisySign, collaboration ensured concentration over $m$. In the next section, we utilize collaboration to obtain optimal dependence on $m$. 

\section{Optimal Dependence on $m$}
\label{sec:optimal_m}

If $\norm{g}_\infty$ or $\norm{g}_2$ is bounded, we can obtain an almost optimal exponential decay with $m$.  %
We provide two schemes that obtain optimal $\ell_\infty$ ( by modifying the sign compressor) and $\ell_2$ error dependence in terms of $m$ and the diameter of the space $B$. The proofs for this section are provided in \Cref{sec:optimal_m_proofs}.

\subsection{HadamardMultiDim}
\label{sec:hadamard}

When the vectors have bounded $\ell_\infty$ norm, instead of obliviously using the sign compressor on every coordinate on every client, one may be able to divide their range and cleverly select bits to encode the most information. We call our algorithm Hadamard scheme, because the binary-search method involved is akin to the rows of a Hadamard-type matrix.
\begin{assumption}[Bounded domain]\label{assump:bdd_grad}
$\norm{g_i}_\infty\leq B,\forall i \in [m]$.
\end{assumption}
This would imply that for any $j\in [d]$, $g_i^{(j)} \in [-B,B], \forall i\in [m]$. Now, consider the $i^{th}$ client and the scalar $g_i^{(j)}$ and assume that we are allowed to encode this using $m$ bits. The best error that we can achieve is $\frac{B}{2^{m-1}}$, by performing a binary search on the range $[-B,B]$ for $g_i^{(j)}$, sending one bit per level of the binary search. However, this scheme is not collaborative. To obtain a collaborative scheme, for some permutation $\rho$ on the set of clients $[m]$, the $i^{th}$ client can perform binary search until level $\rho^{(i)}$ and sends its decision at level $\rho^{(i)}$. In this case, each client sends only $1$ bit per coordinate. To decode  $\tilde{g}^{(j)}$, we take a weighted sum of the signs obtained from different clients weighed by their coefficients $\frac{B}{2^{\rho^{(i)} - 1}}$.  This is the core subroutine (\Cref{alg:hadamard1denc}). The full compression scheme for $d$ coordinates applies this coordinate-wise in~\Cref{alg:hadamardmultidim}. Note that the clients and the server should share the permutation $\rho$ before encoding and decoding, which need not change over different instantiations of the mean estimation problem. 
To understand the core idea of the scheme, consider the case when all vectors $g_i = g$. Then, sending a different level from a different client is equivalent to doing a full binary search to quantize $g$. As long as $g_i$s are close to $g$, we hope that this scheme should give us a good estimate of $g$. Let $\tilde{b}_{i,k}^{(j)}$ denotes the encoding of $g_i^{(j)}$ at level $k$ $\forall i,k\in [m], j\in [d]$. Therefore, if we were to use binary search to encode $g_i^{(j)}$, then $\sum_{k=1}^m \frac{B\tilde{b}_{i,k}^{(j)} }{2^{k-1}}$ would be it's decoded value.

\begin{theorem}[HadamardMultiDim Error]\label{thm:hadamard_error}
Under Assumptions ~\ref{assump:bdd_grad}, the estimation error for Algorithm~\ref{alg:hadamardmultidim} is  
\begin{align}
\E_{\rho\sim \Pi_m}[\norm{\tilde{g} -g}_{\infty}] \leq  \frac{B}{2^{m-1}}+ \min\{\Delta_{\rm Hadamard},\Delta_{\infty,\max}\},
\end{align}
where $\Delta_{\rm Hadamard}\equiv \max_{r\in [d]}\sqrt{\frac{1}{m^2} \underset{1 \leq i\neq j \leq m}{\sum\sum}\sum_{k=1}^m \left(\frac{B(\tilde{b}_{i,k}^{(r)}-\tilde{b}_{j,k}^{(r)})}{2^{k-1}}\right)^2}$ , and $\Delta_{\infty,\max} \equiv \max_{r\in [d],i\in [m]}\abs{g_i^{(r)} - g^{(r)}}$.  
\end{theorem}

As we allow full collaboration between clients, in the worst case, we might have to incur a cost $\Delta_{\infty, \max}$ which is the worst case dissimilarity among clients. However, if client vectors are close, we might end up paying a much lower cost of $\Delta_{\rm Hadamard}$.

\subsection{Sparse regression coding}
\label{sec:sparsereg}
In this part, we extend the coordinate-wise guarantee of the HadamardMultiDim to $\ell_2$ error between $d$-dimensional vectors of bounded $\ell_2$-norm. %
\begin{assumption}[Norm Ball]\label{assump:norm_ball}
$\norm{g_i}_2 \leq B, \forall i \in [m]$.
\end{assumption}

To extend the idea of binary search and full collaboration from HadmardMultiDim, we first need a compression scheme which performs binary search on $d$ dimensional vectors with $\ell_2$ error guarantees. Sparse Regression codes~\cite{venkataramanan_lossy_2014,venkataramanan_lossy_2014-1}, which are known to achieve the optimal rate-distortion tradeoff for a Gaussian source, fit our requirements. Let $A\in \R^{mL\times d}$ be a matrix for some parameter $L>0$, where  $A_k$ denotes the $k$th row of $A$. To compress a single vector $g$ by Sparse Regression codes with the matrix $A$, we use $m\log L$ bits to find the closest vector to $g$ in the first $L$ rows of $A$; say the index of this vector is $\tilde{b}_{1}$. Similar to binary search, we subtract $c_1 A_{\tilde{b}_{1}}$ from $g$, where $c_1$ is given in \eqref{eq:coeff} to obtain an updated $g$. We repeat the process using the next set of $L$ rows. Here, each set of $L$ rows corresponds to a single level of binary search, with the coefficients $c_i$ obtained from Eq~\eqref{eq:coeff} having a decaying exponent.
Although ~\cite{venkataramanan_lossy_2014} requires $A$ to be a random matrix with iid Gaussian entries, if $A$ satisfies the following definition, then we can apply the theoretical guarantees of ~\cite{venkataramanan_lossy_2014}. 

\begin{definition}[$(\delta_1, \delta_2, \Gamma)$-Cover]\label{assump:eps_net}
A matrix $A\in \mathbb{R}^{mL\times d}$ is a $(\delta_1, \delta_2, \Gamma)$-Cover for $\delta_1, \delta_2\in (0,1)$ and $\Gamma > 0$, if for each $k\in [m]$, $\exists \epsilon_k, \gamma_k, \Gamma_k \in \R$, such that 
\begin{align*}
    &\max_{j\in [(k-1)L + 1, kL]} \norm{A_{j}}_2 \leq \sqrt{d}(1+\gamma_k); \quad     \max_{j\neq j'\in [(k-1)L + 1, kL]}\norm{A_j - A_{j'}}_2 \leq \Gamma_k\\
    &\forall v\in \mathbb{S}^{d-1}, \max_{j\in [(k-1)L+1, k L ]} \ip{v}{A_{j}} \geq \sqrt{2\log L}(1+\epsilon_k),\quad\\
    &\sum_{k=1}^m \frac{\abs{\gamma_k}}{m} \leq \delta_1,\quad  \sum_{k=1}^m\frac{\abs{\epsilon_k}}{m}\leq \delta_2,\text{ and }\max_{k\in [m]}\Gamma_k \leq \Gamma .
\end{align*}   
\end{definition}

By the above definition, for each $k\in [m]$, the $L$ consecutive rows of $A$, which correspond to  $\{A_{j}, j\in [(k-1)L+1, kL]\}$, form a $\varepsilon$-net~\cite[Chapter~4.2]{Vershynin_2018} of $\mathbb{S}^{d-1}$ with $\varepsilon = \sqrt{2(1 - \sqrt{\frac{2\log L}{d}})}$. By carefully selecting the parameters in the
proof of ~\citep[Theorem~1]{venkataramanan_lossy_2014}, we can show that this scheme obtains $\ell_2$ error $B\exp(-m)$ as long as $A$ is a $(\delta_1, \delta_2, \Gamma)$-cover. Note that the matrix $A$ is deterministically constructed  with $m$ sets of $\varepsilon$-nets. The choice of Gaussian $A$ in ~\cite{venkataramanan_lossy_2014} can now be understood from the fact that such random Gaussian matrices are in fact $\varepsilon$-nets. The following remark summarizes this.

\begin{remark}[Gaussian $A$]\label{rem:rand_mat}
    If each entry of the matrix $A$ is sampled iid from $\cN(0,1)$ then, for any $\delta_1, \delta_2, \delta_3\in [0,1)$ and $\Gamma^2 = 2d + 4\sqrt{d\log(\frac{mL^2}{\delta_3})} + 4\log(\frac{mL^2}{\delta_3})$, the matrix $A$ is a $(\delta_1, \delta_2, \Gamma)$-Cover with probability $1-\delta_3 - 2m^2Le^{-\frac{d\delta_1^2}{8}} - m\left(\frac{L^{2\delta_2}}{\log L}\right)^{-m}$.
\end{remark}
Note that the expressions for $\delta_1, \delta_2$ can be obtained from ~\cite{venkataramanan_lossy_2014} and the proof for $\Gamma$ is provided in Appendix~\ref{sec:corr_sparsereg_proof}. For $d = \Omega(\log m)$, the probability of error in the above Remark can be made arbitrarily close to $1$ for large $m$.

We now extend the scheme based on sparse regression codes in \cite{venkataramanan_lossy_2014} to our setting of distributed mean estimation, similar to HadamardMultiDim. Each client $i\in [m]$ encodes at level $\rho^{(i)}$ where $\rho$ is a permutation on $[m]$ and the server computes the weighted sum of the encodings from each client with corresponding coefficients $c_{\rho^{(i)}}$. Here, $\tilde{b}_{i,k}$ denotes the encoding of $g_i$ at level $k$ $\forall i,k\in [m]$. If we were to use Sparse Regression codes to encode only $g_i$ then $\sum_{k=1}^m c_k A_{(k-1)L + \tilde{b}_{i,k}}$ would be it's decoded value. The full scheme in described in Algorithm~\ref{alg:sparc_1} and we compute it's $\ell_2$ estimation error in the next theorem. In this scheme, each client only needs to send a single index $\tilde{b}_{i,\rho^{(i)}}$. As $\rho$ is known to the server, this index is in the range $[L]$ and each client uses only $\log L$ bits for communication.

\begin{theorem}[SparseReg Error]\label{thm:sparsereg} Under Assumption~\ref{assump:norm_ball}, %
the following statements hold for $\tilde{g}$, the output of ~\Cref{alg:sparc_1} for any matrix $A\in \R^{mL\times d}$.
\begin{align}
&\E_{\rho \sim \Pi_m}[\norm{g - \tilde{g}}_2^2] = \norm{\bar{g} - g}_{2}^2 + \Delta_{\rm reg},\label{eq:sparsereg_1}\\
&\text{ where, }\Delta_{\rm reg} \equiv \frac{1}{m^2}\underset{i,j\in [m], i\neq j}{\sum\sum}\sum_{k=1}^m c_k^2 \norm{A_{(k-1)L + \tilde{b}_{i,k}} - A_{(k-1)L + \tilde{b}_{j,k}}}_2^2\text{ and }\bar{g} \equiv \frac{1}{m}\sum_{i=1}^m \sum_{k=1}^m c_k A_{(k-1) L + \tilde{b}_{i,k}}.\nonumber
\end{align}
In addition, if the matrix $A$ is a $(\delta_1, \delta_2, \Gamma)$-Cover, then, 
\item \begin{align}
    &\E_{\rho \sim \Pi_m}[\norm{g - \tilde{g}}_2^2] \leq \Lambda + \Delta_{2,\max},\quad \text{and,  }\quad \norm{\bar{g} - g}_2^2 \leq \Lambda, \label{eq:sparsereg_2}\\
    &\text{where,  } \Lambda \equiv B^2\left(1 + \frac{10\log L}{d}e^{\frac{m\log L}{d}}(\delta_1 + \delta_2) \right)^2\left(1 - \frac{2\log L}{d}\right)^{m}.\nonumber\\  
     & \Delta_{\mathrm{reg}}\leq \frac{2B^2 \Gamma^2\log L}{d^2 m^2}\underset{i,j\in [m], i\neq j}{\sum\sum}\sum_{k=1}^m  \left(1-\frac{2\log L}{d}\right)^{k-1} \mathbf{1}(\tilde{b}_{i,k} \neq \tilde{b}_{j,k}). \label{corr:sparsereg}
\end{align}

\begin{remark}[Growth of $\Lambda$ with $m$]
By expanding the squares, we can see that $\Lambda$ can be bounded by 3 terms. 
\begin{align*}
        \Lambda \leq B^2 e^{-\frac{2m\log L}{d}}  +\frac{20 B^2 \log L}{d}(\delta_1 + \delta_2) e^{-\frac{m\log L}{d}}     + \frac{100 B^2 \log^2 L}{d^2} (\delta_1 + \delta_2)^2 .
    \end{align*}
    The first and the second terms decrease exponentially with $m$. The third term is a small constant independent of $m$, depending on $\delta_1,\delta_2 \ll 1$ that can be made arbitrarily small. 
\end{remark}

\end{theorem}

 The proof is provided in Appendix~\ref{sec:sparsereg_proof}. Similar to HadmardMultiDim, the first term in the estimation error, $\Lambda$, has an exponential dependence on $m$ and is obtained from the existing results of Sparse Regression Codes from ~\cite{venkataramanan_lossy_2014}. In terms of $\ell_2$ error this dependence on $m$ is better than all the prior methods. Note that if our matrix $A$ is a $(0,0,\Gamma)$-Cover, i.e., $\delta_1=\delta_2=0$, then $\Lambda$ decreases exponentially with $m$. As mentioned in Remark~\ref{rem:rand_mat}, we can can achieve very small (but nonzero) values for $\delta_1, \delta_2$ with Gaussian matrices. %

By combining Eqs~\eqref{eq:sparsereg_1} and ~\eqref{eq:sparsereg_2}, we can bound the squared $\ell_2$ estimation error by  $\Lambda+ \min\{\Delta_{\rm reg}, \Delta_{2, \max}\}$. Note that this error resembles ~\Cref{thm:hadamard_error}, as it contains an exponentially decreasing term of $m$ along with a minimum of two terms, one of which is the maximum deviation between any client and the other($\Delta_{2,\max}$) and the other is the average difference of encoded values across all pairs of clients ( ($\Delta_{\rm reg}$)). Further, even the dissimilarity term $\Delta_{\rm reg}$ has a similar structure to $\Delta_{\rm Hadamard}$ as it is the pairwise difference between encodings of two different vectors at all levels, where the difference at level $k$ is proportional to $e^{-k}$. If the client vectors are not close to each other, we might incur the worst possible error $\Delta_{2,\max}$, but if they are close, we will pay an average price in terms of $\Delta_{\rm reg}$.

The upper bound on $\Delta_{\rm reg}$ in Eq~\eqref{corr:sparsereg} depends on $A$ only via the encoded values $\{\tilde{b}_{i,k}\}_{i,k\in [m]}$. For this upper bound,  the corresponding terms for each level, here denoted by $k$, decrease geometrically with $k$.  The term $\frac{2B^2 \Gamma^2 \log L}{d^2m^2}$, ensures that $\Delta_{\rm reg}$ is of the correct scale as in the worst case when all encodings are different for all pairs of clients, the summation gives us $\Theta(\frac{d}{2\log L})$ which results in an overall $\Delta_{\rm reg}$ of the scale of $B^2$. From ~\cite{venkataramanan_lossy_2014-1}, for each $k\in [m]$,  all encodings upto and including the $k^{th}$ encoding, i.e., $\{\tilde{b}_{i,k'}\}_{k'\leq k}$, form an estimate of $g_i$ that has error scaling as $B e^{-\frac{\log L (k-1)}{d}}$, which is similar to the error scaling of HadamardMultiDim for a single dimension. %

While  both the  HadmardMultiDim and SparseReg schemes achieve very low communication rate, that comes at the price of $O(m)$ computing in the  \texttt{Encode} step. This higher cost in computing is to be expected when one wants to exploit the full potential of collaborative compression (e.g., \cite{jiang_correlation_2023}, where the \texttt{Decode} step takes $O(m^2)$ time).

\subsection{Benefits of HadmardMultiDim and SparseReg}
\label{sec:example}

 We now provide a example to show that for practical scenarios, the error terms $\Delta_{\mathrm reg}$ and $\Delta_{\mathrm Hadamard}$ are much smaller than their worst case values $\Delta_{\infty, \max}$ and $\Delta_{2,\max}$ and have similar behavior as their average values, $\Delta_2\equiv\frac{1}{m}\sum_{i=1}^m \norm{g_i - g}_2^2$ and $\Delta_{\infty}\Delta_{\infty} \equiv \max_{r\in [d]}\frac{1}{m}\sum_{i=1}^m \abs{g_i^{(r)} - g^{(r)}}$ respectively.  Consider the scenario of Theorem~\ref{thm:hadamard_error} ($\ell_\infty$ error) and set $d=1$. Assume that the first $c$ vectors are $g_1'$
and the remaining $m-c$ vectors  are
$g_2'$, for some constant $c\ll m$. In this case, $\Delta_{\infty, \max} = (1 - \frac{c}{m})\abs{g_1' - g_2'}\approx \abs{g_1' - g_2'}$, while  $\Delta_{\infty} \approx \frac{2c}{m}\abs{g_1' - g_2'}$. In this scenario, if the compressed values $\tilde{b}$
for $g_1'$  and $g_2'$ according to the HadamardMultiDim differ at $k\in \mathcal{K}\subseteq [m]$ levels , then, $\Delta_{\mathrm{Hadamard}}\approx \sqrt{\frac{c}{m} \sum_{k\in \mathcal{K}}(B/2^{k-1})^2}\leq \sqrt{\frac{c B^2}{m} \sum_{k = k^\star}^{\infty}2^{-2(k-1)}}\leq \sqrt{\frac{4c}{3m}} \frac{B}{2^{k^\star-1}}$, where $k^\star = \min_{k\in \mathcal{K}} k$ and we upper bound the sum over the set $\mathcal{K}$ by the sum of a geometric series. \textbf{As  $\Delta_{\mathrm{Hadamard}}$ averages over all machines, it decreases with $m$ similar to $\Delta_2$ and should be much smaller than $\Delta_{\infty, \max}$}. The only case when it is not smaller than $\Delta_{\infty, \max}$ is when
$g_1'$ and $g_2'$ are very close, so that $\Delta_{\infty,\max} = \mathcal{O}(\sqrt{m^{-1}})$, but the first level where they differ ($\min_{k\in \mathcal{K}}k$) is very small. One such example is when the quantized values of $g_1'$
in the set $\mathcal{K}$ sorted by the levels in increasing order are $(+1, -1,-1,-1)$ and that of $g_2'$
are $(-1, +1, +1, +1)$. As the vectors are extremely close in this case, the estimation error with $\Delta_{\infty,\max}$ is not very large. Further, if we assume a distributional assumption on the vectors $g_i$, similar to how we generate Figure~\ref{fig:uniform}, obtaining vectors  where $\Delta_{\mathrm{Hadamard}} > \Delta_{\infty, \max}$, happens with low probability. 
\begin{minipage}{0.48\textwidth}
\begin{algorithm}[H]
\centering
\caption{SparseReg}
\label{alg:sparc_1}
    \begin{algorithmic}
        \STATE  \texttt{\underline{Init()}}
        \STATE Clients and server share $A\in \R^{mL \times d}$, and a permutation $\rho\sim \Pi_m$. 
        \STATE \underline{\texttt{Encode($g_i$)}}
        \STATE $g_i' \gets g_i$
        \FOR{$ j \in [\rho^{(i)}]$}
            \STATE $ \tilde{b}_{i,j} \gets \argmax_{r\in [L]} \ip{A_{(j-1)L + r}}{g_i'} $
            \STATE $g_i' \gets g_i' - c_j A_{(j-1)L + \tilde{b}_{i,j}}$
        \ENDFOR
        \STATE $\tilde{b}_i \gets \tilde{b}_{i,\rho^{(i)}}$
        \RETURN $\tilde{b}_{i}$
        \STATE \texttt{\underline{Decode($\{\tilde{b}_{i}\}_{i\in [m]}$)}}
        \STATE $\tilde{g} \gets \sum_{i\in [m]} c_{\rho^{(i)}} A_{(\rho^{(i)}-1)L + \tilde{b}_i}$
    \end{algorithmic}
\end{algorithm}
    
\end{minipage}%
\hfill
\begin{minipage}{0.49\textwidth}
\begin{align}\label{eq:coeff}
    c_i = B\sqrt{\frac{2\log L}{d^2}\left(1 - \frac{2\log L}{d}\right)^{i-1}}
\end{align}
\vspace{-4mm}
    \begin{algorithm}[H]
    \centering
    \caption{OneBit}
    \label{alg:onebit}
    \begin{algorithmic}
        \STATE \underline{\texttt{Init()}}
        \STATE Clients and server share unit vectors $\{z_i\}_{i\in [m]}$.
        \STATE \underline{\texttt{Encode($g_i$)}}
        \STATE $\tilde{b}_i \gets {\rm sign}(\ip{g_i}{z_i})$
        \RETURN $\tilde{b}_i$
        \STATE \underline{\texttt{Decode($\{\tilde{b}_i\}_{i\in [m]}$)}}
        \STATE $g' \gets \begin{cases}
            &\text{~\citep[Algorithm~1]{pmlr-v206-shen23a}}\,\, \text{(Tech.  I)}\\
            & \frac{1}{m}\sum_{i=1}^{m} z_i \tilde{b}_i\,\, \text{(Tech.  II)}
            \end{cases}$
        \STATE $\tilde{g}\gets g'/\norm{g'}_2$
    \end{algorithmic}
\end{algorithm}
\end{minipage}

Using this example, we can compare the error of our proposed schemes to baselines mentioned in Table~\ref{tab:related_works}. Assume a modification of the above example where each vector is $ d$-dimensional, and all of it's coordinates are either $g_1' \mathbb{1}_d$ or $g_2'$. Consider any $\ell_2$ compressor whose error is either proportional to $\Psi\tilde{B}^2$
or $\Psi \Delta_2$ and it sends $\psi$ bits/client for some constants $\psi, \Psi  >0$.  The $\ell_2$ error is defined as $\E[\norm{\tilde{g} - g}_2^2] $ and the $\ell_\infty$ error is defined as $\E[\norm{\tilde{g} - g}_\infty]$, therefore the corresponding $\ell_\infty$ error of these compressors is $\sqrt{\Psi }\tilde{B}$
or $\sqrt{\Psi}\Delta_2$. Now, consider the example that we just presented with $d>1$ and all coordinates being equal for each vector. Therefore, $\Delta_2 \approx \frac{cd}{m}\abs{g_2' - g_1'}^2$, and plugging this in, the $\ell_2$ error of the schemes is $\sqrt{\Psi }\tilde{B}$
or $\sqrt{\Psi\frac{cd}{m}}\abs{g_2' - g_1'}$. HadamardMultiDim sends $d$ bits/client; therefore, to compare with any of these schemes, we set $\psi=d$. For RandK, this would mean setting $K = \frac{d}{32 + \log d}$. Now, if $\abs{g_1'},\abs{g_2'} \approx B$
but $\abs{g_2' - g_1'} \ll B$, then $\tilde{B} \approx \sqrt{d}B$. Using these approximations, the error of RandK is $\sqrt{(32 + \log d)d}B$  as $\Psi = \frac{d}{K} = 32 + \log d$. This is much larger than the $\ell_\infty$ error of HadamardMultiDim, as the first term is $B\cdot 2^{m-1}$ and the second term $\Delta_{\mathrm{Hadamard}} \approx \sqrt{\frac{c}{m}}\abs{g_2' - g_1'}$. A similar argument holds for all independent compression schemes, as their $\ell_\infty$ error scales as $\tilde{B}$, which in the worst case is $\sqrt{d}B$.

For compressors whose error scales as $\Psi \Delta_2$
(PermK, RandKSpatial, RandKSpatialProj), we obtain the same number of bits/client as HadamardMultiDim scheme by setting $K = \frac{d}{32 + \log d}$.  Consider RandKSpatialProj, where $\Psi = \frac{32 + \log d}{m}$, and the error for our example is $\sqrt{c\frac{(32 + \log d)d}{m^2}}\abs{g_2' - g_1'}$. As long as $d>m$, this error is larger than $\Delta_{\mathrm{Hadamard}}$ by constant terms. A similar argument holds for RandKSpatial and PermK. Additionally, note that the theoretical guarantees for RandKSpatial and RandKSpatialProj do not hold if the correlation is unknown. Without this information, the heuristics they use do not result in theoretical guarantees, and their error might be as bad as RandK. CorrelatedSRQ achieves the lower bound for collaborative compressors for $d=1$, and is based on a coordinate-wise scheme, hence the $\Delta_{\infty}$ in its error guarantees. However, for $d\gg1$, its error scales poorly. For the example described above, $\norm{g_i}_2 \leq \sqrt{d}B$, therefore, the $\ell_\infty$ error for CorrelatedSRQ is
$\sqrt{\frac{1}{m}\min\{\frac{d\Delta_{\infty}^d B}{K},\frac{d^3 B^2}{K^2}\}}$. Even for $K=2$, correlated SRQ requires double the number of bits/client as HadamardMultiDim. The first term of HadamardMultiDim is $B\cdot 2^{m-1}$, which is much smaller than any of these terms, while
$\Delta_{\mathrm{Hadamard}} \approx \sqrt{\frac{m}{c}}\Delta_{\infty}$ for our example. Therefore, as long as
$\left(\frac{m^2 K}{cdB}\right)^{1/(2d-1)} < \Delta_{\infty} <\frac{\sqrt{c}dB}{mK}$, $\Delta_{\mathrm{Hadamard}}$ is smaller than $\ell_\infty$ error of CorrelatedSRQ. The size of this interval for $\Delta_{\infty}$ increases as $d$ increases.

With the above example and analysis, we have specified the exact scenarios when HadamardMultiDim outperforms baselines, and this can be easily extended to SparseReg.

\section{One-bit Schemes}
\label{sec:one_bit}

In this section, our vectors are assumed to belong on the unit sphere $\mathbb{S}^{d-1}$. Further, our goal is to recover the unit vector in the direction of the average vector $    g = (\frac{1}{m}\sum_{i\in [m]}g_i)/\norm{\frac{1}{m}\sum_{i\in [m]}g_i}_2
$. 
\begin{assumption}[Unit vectors]\label{assump:unit_vectors}
 $g_i\in \mathbb{S}^{d-1}, \forall i\in [m]$.
\end{assumption}

Consider the collaborative compressor where each client has sample $z_i \sim {\rm Unif}(\mathbb{S}^{d-1})$ (which are also available to the server apriori). Client $i$ sends the single bit $\tilde{b}_i = {\rm sign}(\ip{g_i}{z_i})$ to the server. To recover $g$, consider the trivial case when all vectors $g_i$s were equal. Then, each $\tilde{b}_i = {\rm sign}(\ip{g}{z_i})$, and to recover $g$, the server needs to learn the halfspace corresponding to $g$ from a set of $m$ labeled datapoints. Applying the same method to when $g_i$s are  not all the same, we can estimate $g$ by solving the following optimization problem. %
\begin{equation}\label{eq:one_bit_opt}
 \min_{\tilde{g}\in \mathbb{S}^{d-1}} \frac{1}{m}\mathbf{1}(\tilde{b_i} \neq {\rm sign}(\ip{z_i}{\tilde{g}}))   .
\end{equation}
Here, $\mathbf{1}(\cdot)$ denotes the indicator function. We can intuitively view ~\eqref{eq:one_bit_opt} as a halfspace learning problem with a groundtruth $g$, but in the presence of noise, as $g_i\neq g$.  %
Learning halfspaces in the presence of noise is hard in general~\cite{hardness}. %
In our setting, if we sample $z_i$ from the intersection of the halfspaces with normal vectors $g$ and $g_i$, then the label is ${\rm sign}(\ip{g}{z_i})$, otherwise, it is $-{\rm sign}(\ip{g}{z_i})$. We can consider this to be under the malicious noise model, wherein a fraction of datapoints are corrupted.

\begin{lemma}[Malicious Noise]\label{lem:label flip}
    If $z_i \sim {\rm Unif}(\mathbb{S}^{d-1})$ and $\tilde{b_i} = {\rm sign}(\ip{z_i}{g_i}),\,\forall i\in [m]$, then, with probability $1-\mathcal{O}(\exp(-m\Delta_{\rm corr}))$, $\zeta$, the fraction of the set of datapoints $\{(z_i, \tilde{b_i})\}_{i\in [m]}$ satisfying ${\rm sign}(\ip{z_i}{g_i})\neq {\rm sign}(\ip{g}{z_i})$ is equal to $ \Theta(\Delta_{\rm corr})$,  where $\Delta_{\rm corr} \triangleq \frac{1}{m\pi}\sum_{i=1}^m \arccos(\ip{g_i}{g})$.
\end{lemma}
The proof of the lemma is provided in ~\Cref{sec:label_flip_proof}. Our methods will use $\Delta_{\rm corr}$ to measure the deviation between clients. For small $\Delta_{\rm corr}$, we obtain better performance. %
If $\ip{g}{g_i} \geq 0,\forall i\in [m]$, then 
\begin{align}\label{rem:delta_corr}
  \cos(\pi\Delta_{\rm corr}) \geq \sqrt{\frac{1}{m} + \frac{2}{m^2}\underset{1\leq i < j \leq m}{\sum\sum}\ip{g_i}{g_j}} . 
\end{align}

As long as the corruption level, $\zeta < \frac{1}{2}$, we can hope to recover the halfspace $g$. We provide two techniques -- Techniques I and II, to recover $g$, thus yielding two corresponding \texttt{Decode} procedures. 

The first decoding procedure (Technique I) is  a linear time algorithm for halfspace learning in the presence of malicious noise~\cite[Theorem~3]{pmlr-v206-shen23a} that provides optimal sample complexity and noise tolerance.  

\begin{theorem}[Error of Technique I]\label{lem:one_bit_optimal_bias}
If $\zeta$ defined in Lemma~\ref{lem:label flip} is less than $\frac{1}{2}$, after running  ~\Cref{alg:onebit} with Technique I,  with probability $1 - \delta - \mathcal{O}(\exp(-m \Delta_{\rm corr}))$, we obtain a hyperplane $\tilde{g}$ %
such that, 
$\ip{\tilde{g}}{g} \geq \cos(\pi(\Delta_{\rm corr} + \frac{d}{m}\mathrm{polylog}(d, \frac{1}{m},\frac{1}{\delta}))$.
\end{theorem}

The algorithm itself is fairly complicated. It assigns weights to different points based on how likely they are to be corrupted. The algorithm proceeds in stages, wherein each stage decreases the weights of the corrupted points and solves the weighted version of ~\eqref{eq:one_bit_opt}. The key technique is to use matrix multiplicative weights update (MMWU) ~\cite{Arora2012TheMW} to yield linear time implementation of both these steps, instead of ~\cite{awasthi_balcan_long} which used polynomial time linear programs for this purpose.

Technique II is the simple average algorithm of \cite{servedio_average}, which obtains suboptimal error guarantees for halfspace learning.

\begin{theorem}[Error of Technique II]\label{lem:average_bias}
If $\zeta$ defined in in Lemma~\ref{lem:label flip} is less than $\frac{1}{2}$, after running ~\Cref{alg:onebit} with Technique II, with probability $1-\delta - \mathcal{O}(\exp(-m\Delta_{\rm corr}))$, we obtain a hyperplane $\tilde{g}$ such that, $ \ip{\tilde{g}}{g} \geq \cos(\pi(\sqrt{d}\Delta_{\rm corr} + \frac{d}{\sqrt{m}}\sqrt{\log(\frac{d}{\delta})}))$.
\end{theorem}

The performance of both techniques improves with decrease in $\Delta_{\rm corr}$. Since we have only $m$ bits to infer a $d$-dimensional vector, we require $m > d$, with Technique II requiring $m > d^2$. If we send $t$ bits per client in OneBit, then the number of samples for the halfspace learning is $mt$, thus obtaining the guarantee in Table~\ref{tab:theoretical_results}. The main benefit of OneBit schemes is their extreme communication efficiency. Existing quantization and sparsification schemes require sending at least $\log K$ or $\log d$, where $K$ is the number of quantization levels. The proofs for this section are provided in Appendix ~\ref{sec:one_bit_proofs}.

\paragraph{Comparison to $\ell_2$ compressor.}
Note that, we can use compressor for $\ell_2$ error to first decode the mean and then normalize it to obtain its unit vector. If such a scheme uses $t$ bits and has $\ell_2$ error either $\Lambda \Delta_2$ or $\Lambda \tilde{B}^2$  then its cosine similarity  $\frac{\ip{g}{\tilde{g}}}{\norm{g'}_2\norm{\tilde{g}}_2} \geq 1 - \frac{\Lambda}{2\norm{g'}_2^2} $ for  $\norm{g'}_2 \approx\norm{\tilde{g}}_2$, where $g'= \frac{1}{m}\sum_{i=1}^m g_i$ and $\tilde{g}$ is the estimate of $g'$. To compare this with OneBit Technique I, we send $\lambda$ bits per client to obtain the same communication budget. The cosine similarity of this scheme is $\cos(\pi(\Delta_{\rm corr} + \frac{d}{tm})) $. We can lower bound this similarity by  $1 - 2\pi^2\Delta_{\rm corr}^2 + 2\pi^2\frac{d^2}{m^2 t^2}$ as $\cos(x) \geq 1 - \frac{x^2}{2}$. Comparing this cosine similarity with that obtained for $\ell_2$-compressor, as long as $2\pi^2 \Delta_{\rm corr}^2 + 2\pi^2\frac{d^2}{m^2\beta^2} < \Lambda$, OneBit Technique I performs better. For any sparsification scheme sending $K$ coordinates, $\Lambda$ is at least $\frac{d}{mK}$. If    we set $t= 32K + K\log d$, OneBit Technique I outerperforms the sparsification scheme as long as $\Delta_{\rm corr}$ is small. 

\section{Experiments}
\label{sec:experiments}
\begin{figure}[t!]
    \centering
        \begin{subfigure}[b]{0.32\textwidth}
    \includegraphics[width=\textwidth]{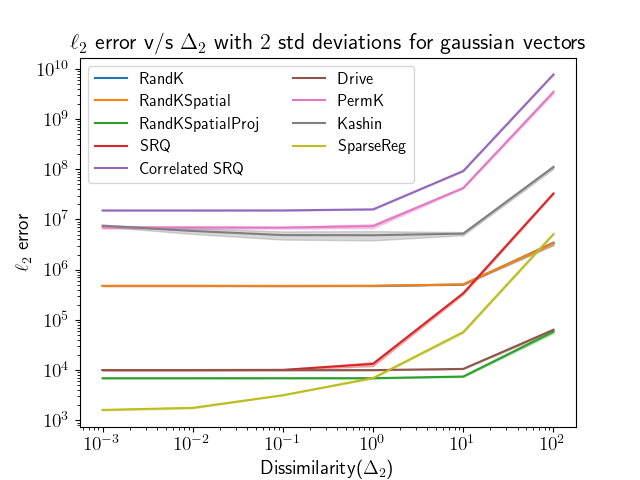}
    \caption{ \small DME with $\ell_2$ error.}
    \label{fig:gaussian}        
    \end{subfigure}
    \hfill
    \begin{subfigure}[b]{0.32\textwidth}
    \includegraphics[width=\textwidth]{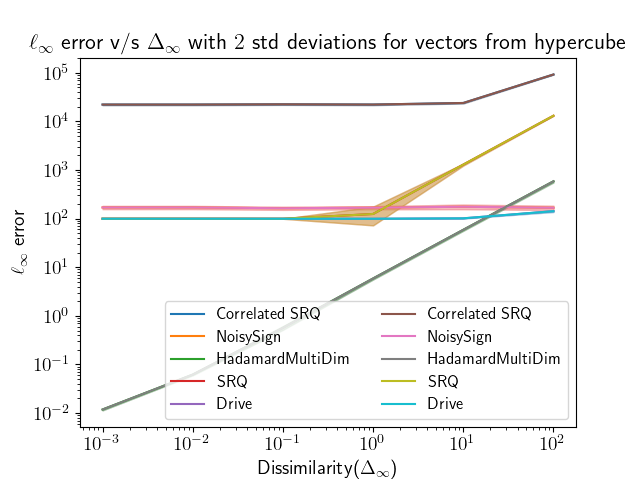}
    \caption{ \small DME with $\ell_\infty$ error.}
    \label{fig:uniform}        
    \end{subfigure}
    \hfill
    \begin{subfigure}[b]{0.32\textwidth}
    \includegraphics[width=\textwidth]{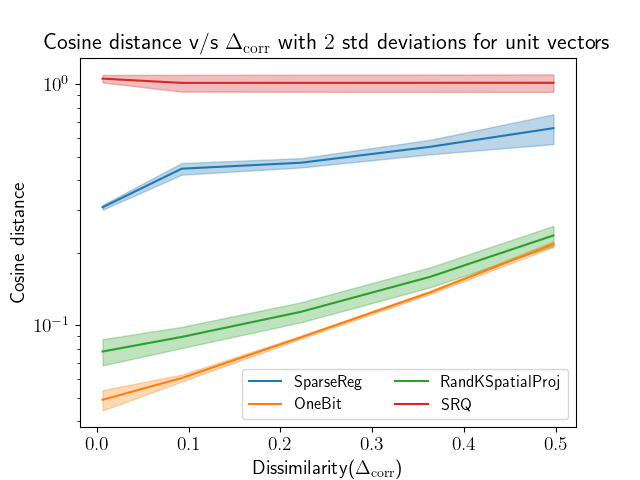}
    \caption{\small DME for cosine distance.}
    \label{fig:unit_norm}        
    \end{subfigure}
\hfill
    \begin{subfigure}[b]{0.32\textwidth}
    \includegraphics[width=\textwidth]{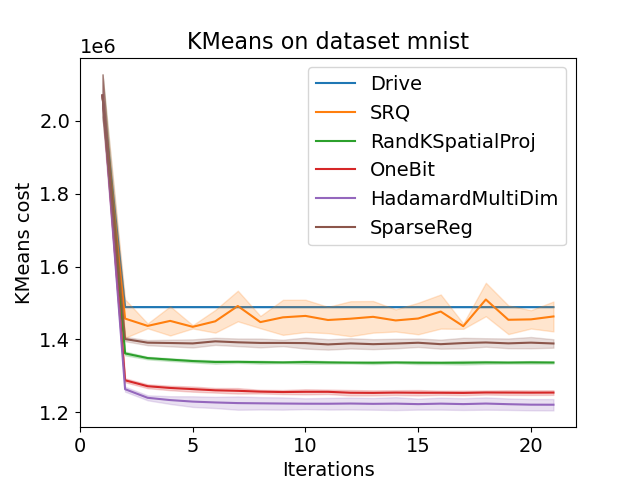}
    \caption{\small KMeans on MNIST.}
    \label{fig:kmeans_mnist}        
    \end{subfigure}
    \hfill
    \begin{subfigure}[b]{0.32\textwidth}
    \includegraphics[width=\textwidth]{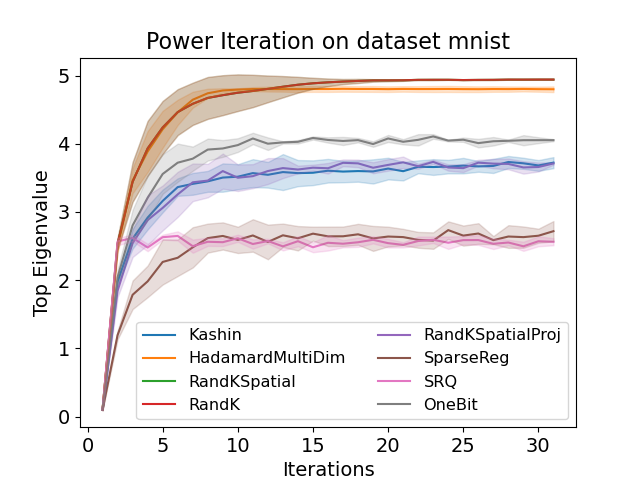}
    \caption{\small Power iteration on MNIST.}
    \label{fig:power_mnist}        
    \end{subfigure}
    \hfill
    \begin{subfigure}[b]{0.32\textwidth}
    \includegraphics[width=\textwidth]{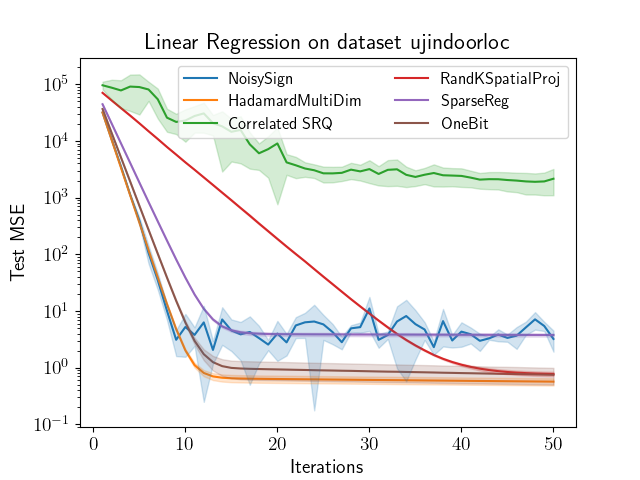}
    \caption{\small Lin. Reg. on UJIndoorLoc.}
    \label{fig:linreg_ujindoorloc}        
    \end{subfigure}
    \begin{subfigure}[b]{0.32\textwidth}
    \includegraphics[width=\textwidth]{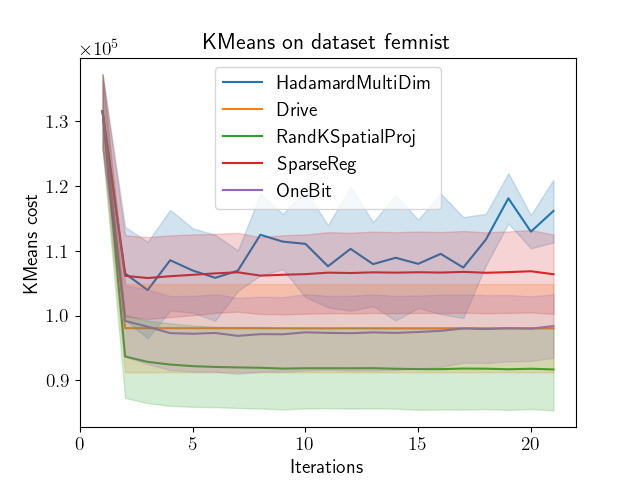}
    \caption{\small KMeans on FEMNIST.}
    \label{fig:kmeans_femnist}        
    \end{subfigure}
    \hfill
    \begin{subfigure}[b]{0.32\textwidth}
    \includegraphics[width=\textwidth]{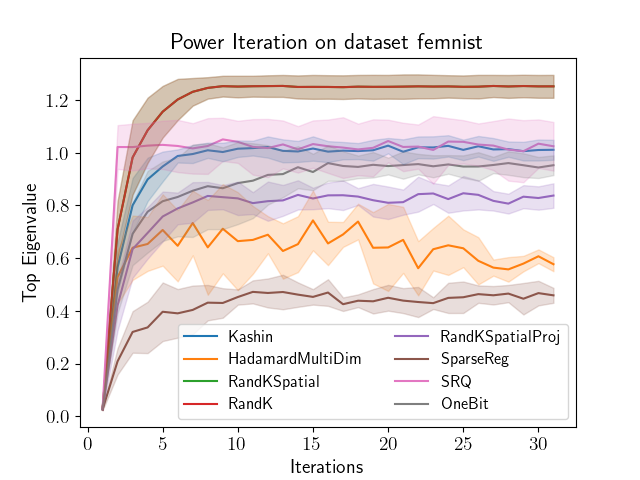}
    \caption{\small Power iteration on FEMNIST.}
    \label{fig:power_femnist}        
    \end{subfigure}
    \hfill
    \begin{subfigure}[b]{0.32\textwidth}
    \includegraphics[width=\textwidth]{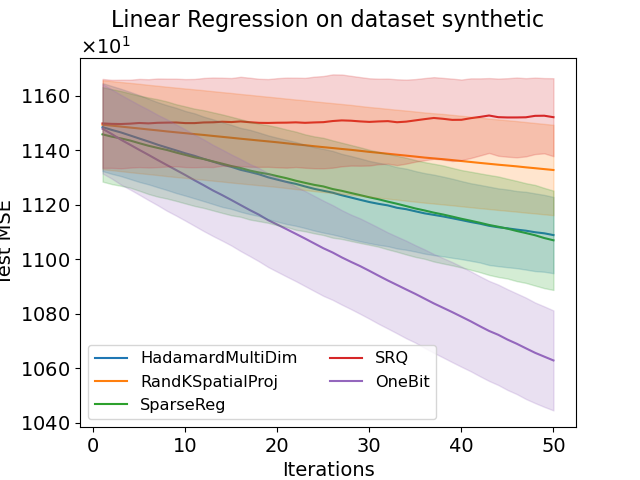}
    \caption{\small Lin. Reg. on Synthetic.}
    \label{fig:linreg_synthetic}        
    \end{subfigure}    
    \caption{\small Performance of DME(Distributed Mean Estimation), KMeans, Power iteration and linear regression for the same communication budget. For each experiment, we report the best compressors. Lin. Reg. refer to Linear Regression. For power iteration, higher top eigenvalue is better. For all other experiments, we report the error, so lower is better.}
    \label{fig:distr_learning}
\end{figure}

\begin{figure}[t!]
    \centering
        \begin{subfigure}[b]{0.4\textwidth}
    \includegraphics[width=\textwidth]{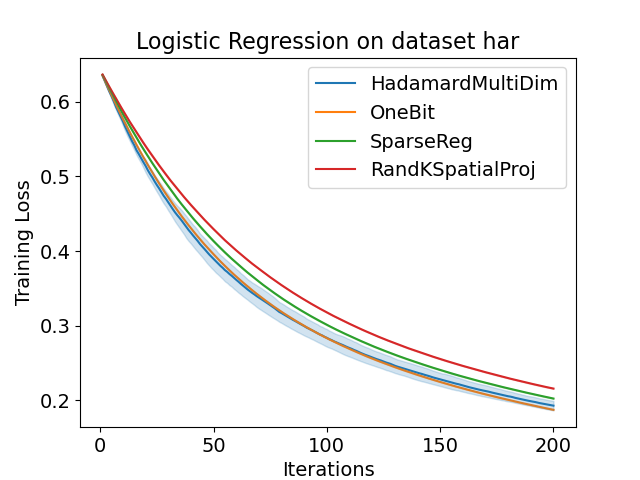}
    \end{subfigure}
    \hfill
    \begin{subfigure}[b]{0.4\textwidth}
    \includegraphics[width=\textwidth]{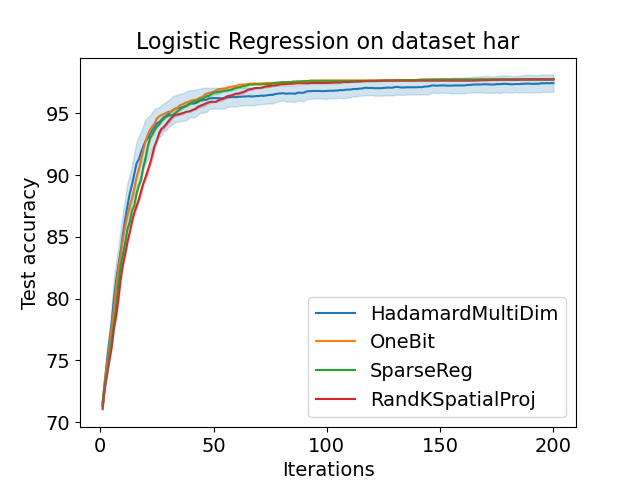}
    \end{subfigure}    
    \caption{Performance of compressors for  Logistic regression on HAR~\cite{har} dataset. Left: Training Logistic loss, Right : Test Accuracy.}
    \label{fig:logreg_har}
\end{figure}

\textbf{Setup.}
To compare the performance of our proposed algorithms, we perform DME for three different distributions which correspond to the three error metrics covered by our schemes -- $\ell_2, \ell_\infty$ and cosine distance. Then, we run our algorithms as the DME subroutine for four different downstream distributed learning tasks -- KMeans, power iteration, linear regression and logistic regression. KMeans and power iteration are run on MNIST~\cite{mnist} and FEMNIST~\cite{leaf} datasets, and we report the KMeans cost and top eigenvalue as the metrics. For linear regression, we run gradient descent on UJIndoorLoc~\cite{misc_ujiindoorloc_310} and a Synthetic mixture of regressions dataset, with low dissimilarity between the mixture components,  and report the test MSE. For logistic regression, we run gradient descent on the HAR~\cite{har} dataset and report the training loss and test accuracy for binary classification. We compare against all baselines in Table~\ref{tab:related_works} for $5$ random seeds and report the methods which perform the best. For DME experiemnts, the shaded corresponds to $2$ standard deviations around the mean, while for all downstream tasks, it corresponds to $1$ standard deviation. The results for KMeans, power iteration and linear regression are reported in Fig~\ref{fig:distr_learning}, while the results for logistic regression are reported in Fig~\ref{fig:logreg_har}. Additional details for our experimental setup are deferred to Appendix~\ref{sec:add_exp}.

\textbf{Results.}
\emph{Distributed Mean Estimation.} From Fig~\ref{fig:gaussian} and ~\ref{fig:uniform}, HadamardMultiDim and SparseReg, whose error is optimal in $m$, obtain the best performance in terms of $\ell_\infty$ and $\ell_2$ error for low dissimilarity. Especially, for HadamardMultiDim in Fig~\ref{fig:uniform}, the gap in $\ell_\infty$ error to next best scheme is very large. NoisySign obtains competitive performance to other baselines as we use a large $\sigma$. The performance of OneBit for cosine distance metric (Fig~\ref{fig:unit_norm}) shows that compressors with $\ell_2$ error guarantees perform poorly in terms of cosine distance. For all collaborative compression schemes, including our proposed schemes, performance degrades as dissmilarity increases. From Fig~\ref{fig:gaussian} and \ref{fig:uniform}, the rate of this decrease is more severe for SparseReg than HadamardMultiDim. For large dissimilarity, HadamardMultiDim and SparseReg can perform worse than certain baselines.

\emph{KMeans and Power iteration.} For MNIST dataset, where dissimilarity is low, HadamardMultiDim performs best for KMeans and close to the best baseline for power iteration (Fig~\ref{fig:kmeans_mnist} and \ref{fig:power_mnist}). Most of our collaborative compression schemes do not perform as well as RandK on FEMNIST, due to  higher client dissimilarity. OneBit is very communication-efficient, so running it for the same communication budget as our baselines ensures that it still remains competitive for KMeans(Fig~\ref{fig:kmeans_femnist}). By increase in confidence intervals for the same baselines from MNIST to FEMNIST, we can see that FEMNIST dataset is much more heterogeneous.

\emph{Linear Regression.}  From Fig~\ref{fig:linreg_ujindoorloc} and\ref{fig:linreg_synthetic}, all collaborative compressors  perform better than independent compressors as UJIndoorLoc and synthetic datasets have low dissimilarity among clients as compared to  FEMNIST. Our schemes can take full advantage of this low dissimilarity, so HadamardMultiDim and SparseReg outperform baselines on both datasets. As the Synthetic dataset has lower dissimilarity than UJIndoorLoc, even the NoisySign performs better than other baselines. Further, OneBit obtains best performance. 

\emph{Logistic Regression.}
From Fig~\ref{fig:logreg_har}, our compressors, Onebit, Sparsereg, and HadamardMultDim, are the best, second, and fourth best compressors, respectively, in terms of both training loss and test accuracy. Further, among the top $4$ best-performing schemes, only one baseline, RandKSpatialProj, comes in third. This shows the benefit of using collaborative compressors. 

\emph{Stability of Our Algorithms.}
Among all our proposed algorithms, OneBit and SparseReg are the most stable with respect to different random seeds followed by HadamardMultiDim and NoisySign. The most stable algorithms among all baselines seems to be RandKSpatialProj. OneBit and Sparsereg are slightly less stable than it. However, all our algorithms except NoisySign are much more stable than most other baselines like SRQ, Correlated SRQ and RandK.

\section{Conclusion}
\label{sec:conclusion}
We proposed four communication-efficient collaborative compression schemes to obtain error guarantees in $\ell_2$-error (SparseReg), $\ell_\infty$-error (NoisySign, HadamardMultiDim) and cosine distance (OneBitAvg). The estimation error of our schemes improves with number of clients, and degrades with increasing dissimilarity between clients. Our schemes are biased and our dissimilarity metrics ($\Delta_{\rm reg}$, $\Delta_{\rm Hadamard}$) depend on the quantization levels. However, these drawbacks can be removed by using existing techniques for converting biased compressors to unbiased ones~\cite{beznosikov_biased_2022}, and adding noise before quantization~\cite{tang_z-signfedavg_2023,chzhen_signsvrg_2023}.
Schemes such as {\em error feedback}~\cite{pmlr-v97-karimireddy19a} reduces the error of independent compressors, and it will be interesting to check if it works for our collaborative compressors.

\paragraph{Acknowledgment.} This work is supported in part by NSF awards 2217058 (EnCORE Inst.), 2112665 (TILOS Inst.), and a grant from Adobe Research. 

\bibliographystyle{tmlr}
\bibliography{references}

\appendix
  \section{Proofs for Section~\ref{sec:noisy_sign}}
\label{sec:noisy_sign_proofs}
\subsection{Proof of ~\Cref{lem:sign_error}}
\label{sec:sign_error_proof} 
As all operations are coordinate-wise, we restrict our focus to only a single dimension $j\in [d]$.

\begin{align*}
    \E_{\xi_i}[\tilde{b}_i^{(j)}] = \Phi_\sigma(g_{i}^{(j)}),\,\forall i \in [m]
\end{align*}
Note that $\Phi_\sigma(t) = \erf(\frac{t}{\sqrt{2}\sigma})$ and $\Phi_\sigma^{-1}(t) = \sqrt{2}\sigma\erf^{-1}(t)$. 
By Hoeffding's inequality for bounded random variables  we have,
    \begin{align*}
        \Pr[\abs{\frac{1}{m}\sum_{i=1}^m(\tilde{b_i}^{(j)} -   \Phi_\sigma(g_{i}^{(j)}))} \geq t] \leq 2e^{-\frac{mt^2}{2}}
    \end{align*}

If we set $t = \sqrt{\frac{2c\log(m)}{m}}$, for some $c>0$ in the above inequality, then with probability $1 - 2m^{-c}$, we have,
\begin{align*}
    \abs{\frac{1}{m}\sum_{i=1}^m (\tilde{b_i}^{(j)} -  \Phi_{
\sigma}(g_{i}^{(j)}))} \leq t
\end{align*}
We can represent $\frac{1}{m}\sum_{i=1}^m \tilde{b}_i = \Phi_\sigma(\tilde{g})$, as $\Phi_\sigma$ is an invertible function. To find the difference between $\tilde{g}$ and $g$, we find the difference $\Phi_\sigma(\tilde{g}) - \Phi_\sigma(g)$. With probability $1-2m^{-c}$, we have,
\begin{align*}
    \abs{\Phi_{\sigma}(\tilde{g}^{(j)}) - \Phi_\sigma(g^{(j)})} \leq \frac{1}{m}\sum_{i=1}^m \abs{\Phi_\sigma(g_i^{(j)}) - \Phi_\sigma(g^{(j)})} + t
\end{align*}
To remove the terms of $\Phi_\sigma$, we can apply the function $\Phi_\sigma^{-1}$ on $\tilde{g}^{(j)}$. As $\Phi_\sigma^{-1}$ is not Lipschitz, we need to perform its Taylor's expansion around $\Phi_\sigma(g^{(j)})$ to account for the linear terms in the error. If $\Delta_\Phi = \frac{1}{m}\sum_{i=1}^m \abs{\Phi_\sigma(g_i^{(j)}) - \Phi_\sigma(g^{(j)})}$, then we obtain,
\begin{align}\label{eq:taylor}
    \abs{\tilde{g}^{(j)} - g^{(j)}} \leq \max_{u \in [\Phi_\sigma(g^{(j)}) - \Delta_\Phi - t,\Phi_\sigma(g^{(j)}) + \Delta_\Phi + t]} \abs{(\Phi_\sigma^{-1})'(u)} (\Delta_\Phi + t)
\end{align}

We now obtain an appropriate upper bound on $(\Phi_\sigma^{-1})'(u)$ as we do not have a closed-form expression for it. We will use the properties of $\erf$ to obtain a suitable bound. First, note that $\Phi_\sigma$ and $\Phi_\sigma^{-1}$ are both odd functions, therefore, $\abs{\Phi^{-1}(u)}  = \abs{\Phi^{-1}(\abs{u})}$, so we consider the bound for $u>0$. Note that $(\Phi^{-1})'(u) = \frac{1}{\Phi'(\Phi^{-1}(u))}$. For $u>0$, we have, 
\begin{align*}
    1 - \erf(u) \leq& e^{-u^2}\\
    \erf(u) \geq & 1 - e^{-u^2}\\
    \erf^{-1}(u) \leq & \sqrt{-\log(1 - u)}\\
    \Phi_\sigma^{-1}(u) = \sqrt{2}\sigma\erf^{-1}(u) \leq & \sigma\sqrt{-2\log(1-u)}\\
    (\Phi_\sigma^{-1})'(u)  = \sqrt{\frac{\pi}{2}} e^{(\Phi_\sigma^{-1}(u))^2/(2\sigma^2)} \leq & \sqrt{\frac{\pi}{2}}e^{-2\log(1 - u)/2} = \sqrt{\frac{\pi}{2}}\frac{1}{1-u}
\end{align*}
For the first step, we use an upper bound on the complementary error function. For the third step, we use the fact that if $f(x) \leq g(x)$, then $f^{-1}(y) \geq g^{-1}(y)$.

Using the following upper bound in Eq~\eqref{eq:taylor}, we obtain,
\begin{align*}
    \abs{\tilde{g}^{(j)} - g^{(j)}} \leq& \max_{u \in [\Phi_\sigma(g^{(j)}) - \Delta_\Phi - t,\Phi_\sigma(g^{(j)}) + \Delta_\Phi + t]} 
\sqrt{\frac{\pi}{2}}\frac{\Delta_\Phi + t}{1- \abs{u}}\\
    \leq & \sqrt{\frac{\pi}{2}}\frac{\Delta_\Phi  + t}{1 - \max\{\abs{\Phi_\sigma(g^{(j)}) - \Delta_\Phi - t},\abs{\Phi_\sigma(g^{(j)}) + \Delta_\Phi + t}\}}
\end{align*}
We use $\max\{\abs{\Phi_\sigma(g^{(j)}) - \Delta_\Phi - t},\abs{\Phi_\sigma(g^{(j)}) + \Delta_\Phi + t}\} \leq \Phi_\sigma(\abs{g^{(j)}}) + \Delta_\Phi + t$, as $\Phi_\sigma$ is an increasing odd function.
\begin{align*}
    \abs{\tilde{g}^{(j)} - g^{(j)}} \leq \sqrt{\frac{\pi}{2}}\left(\left(1 - \frac{\Delta_\Phi +t }{1  - \Phi_{\sigma}(\abs{g^{(j)}})}\right)^{-1} - 1\right)
\end{align*}

We extend the bound to $d$ dimensions by taking a union bound, yielding a probability of error $2dm^{-c}$.
\subsection{Proof of ~\Cref{corr:noisy_sign}}
First, note that the term of $(1+x)^{-1} - 1 = \frac{x}{1-x}\leq \mathcal{O}(x)$ when $x\leq c$ for some constant $c\in [0,1]$. The conditions required for this are $\Delta_\Phi$ to be small and both $m$ and $\alpha(\norm{g}_\infty)$ to be large. Note that $\alpha(\norm{g}_\infty) =  1 - \Phi_{\sigma}(\norm{g}_\infty)$, so it is small when the ratio $\frac{\norm{g}_\infty}{\sigma}$ is small. The bound on estimation error after this step is the following.
\begin{align*}
    \norm{\tilde{g} - g}_\infty = \mathcal{O}\left(\frac{\Delta_\Phi + \sqrt{\frac{\log m}{m}}}{\alpha(\norm{g}_\infty)}\right)
\end{align*}
Now, we lower bound the term $\alpha(\norm{g}_\infty)$. Note that $\alpha(x) = 1 - \Phi_\sigma(x) = \frac{1}{2}(1 - \erf(\frac{x}{\sqrt{2}\sigma}))$ for any $x\in \R_{+}$ where $\erf$ is the error function. Here, $1 - \erf(x)$ is the complementary error function for any $x\geq 0$ and it is lower bounded by $\Omega(e^{-\beta' x^2})$ for $\beta'>1$~\cite{comp_erf_lb}. Plugging in this lower bound, we complete the proof for a $\beta = \beta'/2 > \frac{1}{2}$.

\subsection{Proof of ~\Cref{rem:noisy_sign_delta}}
\label{sec:noisy_sign_delta_proof}
The proof follows from using the triangle inequality and a Taylor's expansion for each $\Phi_\sigma(g_i^{(j)})$ around $g^{(j)}$. Note that, for some $u_i^{(j)}$ between $g^{(j)}$  and $g_i^{(j)}$, we have,
\begin{align*}
    &\Phi_\sigma(g_i^{(j)}) = \Phi_\sigma(g^{(j)}) + \sqrt{\frac{2}{\pi}}\frac{(g^{(j)} - g_i^{(j)})e^{-\frac{(u_i^{(j)})^2}{2\sigma^2})}}{\sigma}\\
    &\abs{\Phi_\sigma(g_i^{(j)}) - \Phi_\sigma(g^{(j)})} \leq  \sqrt{\frac{2}{\pi}}\frac{\abs{g^{(j)} - g_i^{(j)}}}{\sigma}
\end{align*}
We use the fact that $e^{-\frac{(u_i^{(j)})^2}{2\sigma^2}} \leq 1$.
By using triangle inequality for any coordinate $j\in [m]$, we obtain,
\begin{align*}
    \Delta_\Phi \leq& \max_{j\in [d]}\frac{1}{m}\sum_{i\in [m]}\abs{\Phi_\sigma(g_i^{(j)}) - \Phi_\sigma(g^{(j)})} \leq \frac{1}{m}\sum_{i\in [m]}\max_{j\in [d]}\abs{\Phi_\sigma(g_i^{(j)}) - \Phi_\sigma(g^{(j)})}\\
    \leq& \sqrt{\frac{2}{\pi}} \frac{1}{m}\sum_{i\in [m]}\max_{j\in [d]}\frac{\abs{g^{(j)} - g_i^{(j)}}}{\sigma}    \leq \sqrt{\frac{2}{\pi}} \frac{1}{m}\sum_{i\in [m]}\frac{\norm{g - g_i}_\infty}{\sigma}
\end{align*}

\section{Proofs for ~\Cref{sec:optimal_m}}
\label{sec:optimal_m_proofs}

\subsection{Proof of ~\Cref{thm:hadamard_error}}
\label{sec:hadamard_error_proof}
To prove this Lemma, we first consider a single dimension $j\in [d]$. Note that the collaborative compression scheme has precision $\frac{B}{2^{m-1}}$. Since, for each client $g_{i}^{(j)}$ lies in the range $[g^{(j)} - \Delta_{\max},g^{(j)} + \Delta_{\max}]$, therefore, we must decode one of the elements which has error atmost $\Delta_{\max} + \frac{B}{2^{m-1}}$. Since this is true for each coordinate individually, therefore, it holds for the $\ell_\infty$ norm.

Consider a single dimension $j\in [d]$. Let $g_{i}^{(j)}$ be the $j^{th}$ coordinate of $g_i$ and $\rho_j$ be the permutation selected for the coordinate $j$. We omit $j$ from  $g_{i}^{(j)}$ and $\rho_j$ to simplify the notation. Let $\tilde{b}_{i,p}$ be the estimate of $g_i$ after decoding it for $p$ levels where $p\in [m]$. Therefore, the estimator $\tilde{g} = \sum_{i=1}^m \frac{\tilde{b}_{i,\rho_i}B}{2^{\rho_i-1}}$.  Let $\tilde{g}_i = \sum_{k=1}^m \frac{\tilde{b}_{i,k}B}{2^{k-1}}$ be the decoded value of $g_i$ till level $m$ and $\bar{g} = \frac{1}{m}\sum_{i=1}^m \tilde{g}_i = \sum_{k=1}^m \frac{\bar{b}_{k}B}{2^{k-1}}$, where $\bar{b}_k = \frac{1}{m}\sum_{i=1}^m \tilde{b}_{i,k}$.

We compute the expected error for coordinate $j$, where the expectation is wrt the permutation $\rho_j$. Note that $\E_{\rho \sim \Pi_m}[\tilde{g}_i] = \bar{g}$.
\begin{align*}
    \E_{\rho \sim \Pi_m}[\abs{g - \tilde{g}}]  &= \sqrt{(\E_{\rho \sim \Pi_m}[\abs{g - \tilde{g}}])^2} \leq \sqrt{\E_{\rho \sim \Pi_m}\abs{g - \tilde{g}}^2} \leq \sqrt{\E_{\rho \sim \Pi_m}\abs{\tilde{g} - \bar{g}}^2 + \abs{g - \bar{g}}^2}\\
    &\leq \sqrt{\E_{\rho \sim \Pi_m}\abs{\tilde{g} - \bar{g}}^2} + \abs{g - \bar{g}} \leq \frac{1}{m}\sum_{i=1}^m \abs{g_i - \tilde{g}_i} + \sqrt{\E_{\rho \sim \Pi_m}\abs{\tilde{g} - \bar{g}}^2}\\
    &\leq \frac{B}{2^{m-1}} + \sqrt{\E_{\rho \sim \Pi_m}\abs{\tilde{g} - \bar{g}}^2}
\end{align*}
We use Jensen's inequality for the first inequality. For the second inequality, we use bias-variance decomposition for the random variable $\tilde{g}$, where the first term is its variance, and the second term is its bias wrt the term $g$. We then use $\sqrt{a+b}\leq \sqrt{a} + \sqrt{b}$ for any $a,b \geq 0$. To handle the term $\abs{g - \bar{g}}$, we expand both terms as a summation over $m$ clients, followed by a triangle inequality. As each estimator $\tilde{g}_i$ is at least $\frac{B}{2^{m-1}}$ away from $g_i$, each term in the difference $\abs{g_i - \tilde{g}_i}$ has the upperbound $\frac{B}{2^{m-1}}$.

We now bound the variance term separately. Note that
\begin{align*}
    \E_{\rho \sim \Pi_m}\abs{\tilde{g} - \bar{g}}^2 = \E_{\rho \sim \Pi_m}\abs{\tilde{g}}^2  - \bar{g}^2 
\end{align*}
We first evaluate the second moment $\E_{\rho \sim \Pi_m}\abs{\tilde{g}}^2$.
\begin{align*}
    \E_{\rho \sim \Pi_m}\abs{\tilde{g}}^2  &= \E_{\rho \sim \Pi_m}\left\lvert\sum_{i=1}^m \frac{\tilde{b}_{i,\rho_i}}{2^{\rho_i-1}}\right\rvert^2 =     
    \sum_{i=1}^m \E_{\rho \sim \Pi_m}\left[\frac{\tilde{b}_{i,\rho_i}^2]B^2}{2^{2\rho_i-2}}\right] + B^2\underset{1 \leq i\neq j \leq m}{\sum\sum}\E_{\rho \sim \Pi_m}\left[\frac{\tilde{b}_{i, \rho_i}}{2^{\rho_i-1}}\frac{\tilde{b}_{j, \rho_j}}{2^{\rho_j - 1}}\right]\\
    &= \sum_{k=1}^m \frac{B^2}{2^{2k - 2}} + B^2\underset{1 \leq i\neq j \leq m}{\sum\sum}\E_{\rho_i}\left[\E_{\rho \sim \Pi_m}\left[\frac{\tilde{b}_{i, \rho_i}}{2^{\rho_i-1}}\frac{\tilde{b}_{l, \rho_j}}{2^{\rho_j - 1}}| \rho_i\right]\right]\\
    &= \sum_{k=1}^m \frac{B^2}{2^{2k - 2}} + B^2\underset{1 \leq i\neq j \leq m}{\sum\sum}\E_{\rho_i}\left[\frac{\tilde{b}_{i, \rho_i}}{2^{\rho_i-1}}\frac{1}{m-1}\sum_{l=1, l\neq \rho_i}^{m} \frac{\tilde{b}_{j,l}}{2^{l-1}}\right]\\
    &= \sum_{k=1}^m \frac{B^2}{2^{2k - 2}} + \frac{B^2}{m(m-1)}\underset{1 \leq i\neq j \leq m}{\sum\sum}\sum_{k=1}^m\left[\frac{\tilde{b}_{i, k}}{2^{k-1}}\sum_{l=1, l\neq k}^{m} \frac{\tilde{b}_{j,l}}{2^{l-1}}\right]\\
    &= \sum_{k=1}^m \frac{B^2}{2^{2k - 2}} + \frac{1}{m(m-1)}\underset{1 \leq i\neq j \leq m}{\sum\sum}\left(\sum_{k=1}^m\frac{\tilde{b}_{i, k}B}{2^{k-1}}\right)\left(\sum_{l=1}^{m} \frac{\tilde{b}_{j,l} B}{2^{l-1}}\right)\\
    & \quad - \frac{1}{m(m-1)}\underset{1 \leq i\neq j \leq m}{\sum\sum} \sum_{k=1}^m \frac{B^2 \tilde{b}_{i,k}\tilde{b}_{j,k}}{2^{2k-2}}\\
    &= \sum_{k=1}^m \frac{B^2}{2^{2k - 2}} + \frac{1}{m(m-1)}\underset{1 \leq i\neq j \leq m}{\sum\sum}\tilde{g}_i \tilde{g}_j - \frac{1}{m(m-1)}\underset{1 \leq i\neq j \leq m}{\sum\sum} \sum_{k=1}^m \frac{B^2 \tilde{b}_{i,k}\tilde{b}_{j,k}}{2^{2k-2}}\\
    &=\frac{m^2\abs{\bar{g}}^2 - \sum_{i=1}^m \abs{\tilde{g}_i}^2}{m(m-1)} 
    + \frac{1}{m(m-1)} \underset{1 \leq i\neq j \leq m}{\sum\sum}\sum_{k=1}^m \frac{B^2(\abs{\tilde{b}_{i,k}}^2 + \abs{\tilde{b_{j,k}}}^2 - 2\tilde{b}_{i,k}\tilde{b}_{j,k})}{2^{2k-1}}\\
    &= \frac{m}{m-1}\abs{\bar{g}}^2 - \frac{\sum_{i=1}^m \abs{\tilde{g}_i}^2}{m(m-1)}  + \frac{1}{2m(m-1)} \underset{1 \leq i\neq j \leq m}{\sum\sum}\sum_{k=1}^m \left(\frac{B(\tilde{b}_{i,k}-\tilde{b}_{j,k})}{2^{k-1}}\right)^2
\end{align*}
Note that we expand the square of the sum of terms where $\tilde{b}_{i,j}^2 = 1$. For the second term, we use the law of total expectation by conditioning on the value of $\rho_i$. To evaluate the inner expectation, we note that  $\rho_j$ can take any value other than that of $\rho_i$ with equal probability. To evaluate the outer expectation, note that $\rho_i$ can take any value in $[m]$ with equal probability. In the fourth equation, we subtract the term where $l=k$. Then, we can factorize the remaining terms to obtain $\tilde{g}_i$ and $\tilde{g}_j$. Note that the sum of the product terms $\tilde{g}_i\tilde{g}_j$ can be expressed as $\abs{\sum_{i=1}^m \tilde{g}_i}^2$, with the square terms subtracted. Further, we express the term $\frac{B^2}{2^{2k-2}} = \underset{1 \leq i\neq j \leq m}{\sum\sum} \frac{B^2(\abs{\tilde{b}_{i,k}}^2 +\abs{\tilde{b}_{j,k}}^2}{2^{2k-1}}$ as $\abs{\tilde{b}_{i,k}}^2 = 1$. Finally, we complete the squares for each term $k$.

Using the above value of second moment $\E_{\rho \sim \Pi_m}\abs{\tilde{g}}^2$, we can compute the variance,
\begin{align*}
    \E_{\rho \sim \Pi_m}\abs{\tilde{g} - \bar{g}}^2  &= \E_{\rho \sim \Pi_m}\abs{\tilde{g}}^2 - \abs{\bar{g}}^2 = \frac{\abs{\bar{g}}^2 - \frac{1}{m}\sum_{i=1}^m \abs{\tilde{g}_i}^2}{m-1} + \frac{1}{2m(m-1)} \underset{1 \leq i\neq j \leq m}{\sum\sum}\sum_{k=1}^m \left(\frac{B(\tilde{b}_{i,k}-\tilde{b}_{j,k})}{2^{k-1}}\right)^2\\
    &=\frac{1}{2m^2} \underset{1 \leq i\neq j \leq m}{\sum\sum}\sum_{k=1}^m \left(\frac{B(\tilde{b}_{i,k}-\tilde{b}_{j,k})}{2^{k-1}}\right)^2 
\end{align*}
We use $\bar{g}^2 \leq \frac{1}{m}\sum_{i=1}^m \abs{\tilde{g}_i}^2 = \frac{1}{2 m^2}\underset{1 \leq i \neq j \leq m }{\sum\sum} (\tilde{g}_i - \tilde{g}_j)^2 \geq \frac{1}{2 m^2}\underset{1 \leq i \neq j \leq m }{\sum\sum} \sum_{k=1}^m\left(\frac{B(\tilde{b}_{i,k}-\tilde{b}_{j,k})}{2^{k-1}}\right)^2$.

To simplify this bound, we need to incorporate the difference in the actual gradient vectors. For this purpose, we try to bound the differences $\abs{\tilde{b}_{i,k} - \tilde{b}_{j,k}}$ in terms of $\Delta_{ij} \triangleq \abs{g_i - g_i}$. If $\Delta_{ij} = \abs{g_i - g_j}$, then $\tilde{b}_{i,k} = \tilde{b}_{j,k}, \forall k \geq \log\left(\frac{B}{\Delta_{ij}}\right) $.

\subsection{Proof for ~\Cref{thm:sparsereg}}
\label{sec:sparsereg_proof}
We first show how to obtain the coefficients $c_i$ and the bound on the term $\Lambda$ in Eq~\eqref{eq:sparsereg_2}. To obtain the coefficients $c_i$, we replace set $L=m,n=d, R=\frac{m\log L}{d}$ and $\sigma^2 = \frac{B^2}{d}$ in ~\cite[Eq~2]{venkataramanan_lossy_2014}. Note that the terms $\delta_1$ and $\delta_2$ can be directly applied in ~\cite[Section~V]{venkataramanan_lossy_2014} with the same values of $\epsilon_k$ and $\gamma_k$.  Note that the bound on $\Lambda$ in ~\cite[Theorem~1]{venkataramanan_lossy_2014} is for gaussian sources with a fixed variance, however, their proof in ~\cite[Section V]{venkataramanan_lossy_2014} only uses the fact that the squared $\ell_2$ norm of the source is bounded, which is exactly our case from Assumption~\ref{assump:norm_ball}.

The proof of Eq~\eqref{eq:sparsereg_2} is same as Theorem~\ref{thm:hadamard_error} for a single dimension, with the coefficients $\frac{B}{2^{j-1}}$ replaced by $c_j$ and $\tilde{b}_{i,k}^{(r)}$ replaced by $A_{(k-1)L + \tilde{b}_{i,k}}$. Following ~\Cref{sec:hadamard_error_proof}, we can write down the $\ell_2$ error.
\begin{align*}
    \E_{\rho \sim \Pi_m}[\norm{\tilde{g} - g}_2^2] &= \E_{\rho \sim \Pi_m}[\norm{g - \E_{\rho \sim \Pi_m}[\tilde{g}]}_2^2] +  \E_{\rho \sim \Pi_m}[\norm{\tilde{g} - \E_{\rho \sim \Pi_m}[\tilde{g}]}_2^2]
\end{align*}
$\E[\tilde{g}] = \bar{g} = \frac{1}{m}\sum_{i=1}^m \bar{g}_i$, where $\bar{g}_i = \sum_{j=1}^m c_{j}  A_{(j-1)L + \tilde{b}_{i,j}}$. By triangle inequality, the first term is $\frac{1}{m}\sum_{i=1}^m\norm{g_i - \bar{g}_i}_2^2$, which is bounded individually by $\Lambda = B^2(1 + \frac{10\log L}{d}e^{\frac{m\log L}{d}}(\delta_1 + \delta_2) )^2\left(1 - \frac{2\log L}{d}\right)^{m}$ by setting  $L=m,n=d,R=\frac{m\log L}{d}, \sigma^2 = \frac{B^2}{d}$ and $\delta_0=0$ in ~\cite[Theorem~1]{venkataramanan_lossy_2014}. 

For the second term, we need to bound $\E[\norm{\tilde{g}}_2^2]$.
\begin{align*}
    \E[\norm{\tilde{g}}_2^2] &= \frac{1}{m}\sum_{i=1}^m \sum_{j=1}^m c_i^2 \norm{A_{(j-1)L +\tilde{b}_{i,j}}}_2^2 \\
&\quad + \underset{1 \leq i\neq j \leq m}{\sum\sum}\E_{\rho \sim \Pi_m}\left[c_{\pi(i)}c_{\pi(j)}\ip{A_{(\pi(i)-1)L + \tilde{b}_{i,\pi(i)}}}{A_{(\pi(j)-1)L + \tilde{b}_{j,\pi(j)}}}\right]\\
&= \frac{1}{m}\sum_{i=1}^m \sum_{j=1}^m c_i^2 \norm{A_{(j-1)L +\tilde{b}_{i,j}}}_2^2 \\
&\quad + 
\frac{1}{m(m-1)}\underset{1 \leq i\neq j \leq m}{\sum\sum}\E_{\rho \sim \Pi_m}\left[c_{\pi(i)}c_{\pi(j)}\ip{A_{(\pi(i)-1)L + \tilde{b}_{i,\pi(i)}}}{A_{(\pi(j)-1)L + \tilde{b}_{j,\pi(j)}}}\right]\\
    &=\frac{m^2\norm{\bar{g}}_2^2 - \sum_{i=1}^m \norm{\tilde{g}_i}_2^2}{m(m-1)} 
    + \frac{1}{m(m-1)} \underset{1 \leq i\neq j \leq m}{\sum\sum}\sum_{k=1}^m c_k^2 \norm{ A_{(k-1)L + \tilde{b}_{j,k}} - A_{(k-1)L + \tilde{b}_{i,k}}}_2^2
\end{align*}
The remainder of the proof follows proof of Theorem~\ref{thm:hadamard_error} with $\abs{\cdot}^2$ replaced by $\norm{\cdot}_2^2$.

For the first equation in Eq~\eqref{eq:sparsereg_2}, since the collaborative compression scheme has precision $\Lambda$, for each client, the vector $g_i$ lies in a ball around $g$ of radius $\Delta_{2,\max} = \max_{i\in [m]} \norm{g_i - g}_2$.  Therefore, if we use sparse regression codes to individually encode each vector $g_i$, it's decoded value $\tilde{g}_i$ must lie either in the same ball or be at most $\Lambda$ $\ell_2$ distance away from some vector in this ball. Since this set is convex, and the vector $\bar{g}$ is also obtained as a convex combination of vectors inside this ball or at most $\Lambda$ away from it, the vector $\bar{g}$ also lies inside this ball or is at most $\Lambda$ away from it.

This proves Eq~\eqref{eq:sparsereg_2}.

\subsection{Proof of ~\Cref{corr:sparsereg} and ~\Cref{rem:rand_mat}}
\label{sec:corr_sparsereg_proof}

To bound $\Delta_{reg}$, we need to bound the terms of the form $\norm{A_{(k-1)L + \tilde{b}_{i,k}}-A_{(k-1)L + \tilde{b}_{j,k}}}_2^2$ for clients $i\neq j\in [m]$ and at level $k\in [m]$. Note that if $\tilde{b}_{j,k} = \tilde{b}_{i,k}$, then this difference is $0$. If $\tilde{b}_{j,k} \neq \tilde{b}_{i,k}$, then this is $\Gamma^2$ from the definition of $(\delta_1, \delta_2, \Gamma)$-Cover. This results in the following upper bound for $\Delta_{reg}$.
\begin{align*}
    \Delta_{reg} \leq& \frac{\Gamma^2}{m} \underset{i,j\in [m], i\neq j}{\sum\sum}\sum_{k=1}^m  c_k^2 \cdot\mathbf{1}(\tilde{b}_{i,k} \neq \tilde{b}_{j,k})    = \frac{2\Gamma^2 B^2 \log L}{d^2 m} \underset{i,j\in [m], i\neq j}{\sum\sum}\sum_{k=1}^m  \left(1-\frac{2\log L}{d}\right)^{k-1} \mathbf{1}(\tilde{b}_{i,k} \neq \tilde{b}_{j,k})\\
    = &\frac{2\Gamma^2 B^2 \log L}{d m^2} \underset{i,j\in [m], i\neq j}{\sum\sum}\sum_{k=1}^m  \left(1-\frac{2\log L}{d}\right)^{k-1} \mathbf{1}(\tilde{b}_{i,k} \neq \tilde{b}_{j,k})
\end{align*}
We use the value of $c_k$ in the second inequality and then simplify the terms.

To prove the bound of $\Gamma$ for the case of Gaussian $A$ in Remark~\ref{rem:rand_mat}, note that $\Gamma$ is an upper bound on the the squared $\ell_2$ norm of the difference of two gaussian random vectors. For any $r_1, r_2\in [L]$ with $r_1\neq r_2$, $A_{(k-1)L + r_1} - A_{(k-1)L + r_2} \sim \cN(0, 2\I_d)$ as each vector is $d$-dimensional, they are sampled independently, and each element in each vector is sampled from a unit normal. Therefore, $\frac{1}{2}\norm{A_{(k-1)L + r_1} - A_{(k-1)L + r_2}}_2^2$ is a $\chi_d^2$ random variable. Note that the scaling of $\frac{1}{2}$ is to normalize the covariance to $\I_d$. By concentration of $\chi_d^2$ random variables from ~\cite{Laurent2000}, we know that, with probability $1-\delta$,
\begin{align*}
    \norm{A_{(k-1)L + r_1} - A_{(k-1)L + r_2}}_2^2 \leq 2 d + 4\sqrt{d\log(1/\delta)} + 4\log(1/\delta)
\end{align*}
Since $r_1, r_2\in [L]$, we can take a union bound over all $\frac{L(L-1)}{2}$ unique pairs of $(r_1, r_2)$. Further, we can take a union bound over all levels $k\in [m]$. These would set the probability of error to $\frac{mL(L-1)\delta}{2}$. Setting $\delta_3$ to this value, we obtain $\delta = \frac{2\delta_3}{mL(L-1)}$ and the following bound $\forall k\in [m], r_1\neq r_2\in [L]$,
\begin{align*}
    \norm{A_{(k-1)L + r_1} - A_{(k-1)L + r_2}}_2^2 \leq 2d + 4\sqrt{d\log(\frac{mL^2}{\delta_3})} + 4\log(\frac{mL^2}{\delta_3}) \triangleq \Gamma^2
\end{align*}
We upper bound $\frac{L(L-1)}{2}$ by $L^2$.

\subsection{Improving HadamardMultiDim and SparseReg with Repetitions}
Let $\tilde{g}_R = \frac{1}{R}\sum_{r=1}^R \tilde{g}(r)$ be estimator after $R$ repetitions where $\tilde{g}(r)$ is the estimator after $1$ repetition of either HadamardMultiDim or SparseReg. To compute the error for both ~\Cref{thm:hadamard_error} and ~\Cref{thm:sparsereg}, we use a bias-variance decomposition of the squared estimation error. Consider SparseReg where we bound $\E_{\rho \sim \Pi_m}[\norm{\tilde{g} - g}_2^2]$. Note that even with the new definition of $\tilde{g}$, it's mean wrt $\rho$ remains the same as $\E_{\rho \sim \Pi_m}[\tilde{g}] = \frac{1}{R}\sum_{r=1}^R \E_{\rho \sim \Pi_m}[\tilde{g}(r)] = \frac{1}{R}\sum_{r=1}^R \bar{g} = \bar{g}$. This contributes to the term with exponential dependence on $m$ and it remains the same. Note that the variance term, however, decreases with $R$.
\begin{align*}
    \E_{\rho \sim \Pi_m}[\norm{\tilde{g} - \bar{g}}_2^2] = \frac{1}{R^2} \sum_{r=1}^R \E_{\rho \sim \Pi_m}[\norm{\tilde{g}(r) - \bar{g}}_2^2] = \frac{1}{R}\E_{\rho \sim \Pi_m}[\norm{\tilde{g}(1) - \bar{g}}_2^2] = \frac{\Delta_{\rm reg}}{R}
\end{align*}
Similarly for HadamardMultiDim, the term $\Delta_{\rm Hadamard}$ is replaced by $\frac{\Delta_{\rm Hadamard}}{\sqrt{R}}$, as we take square root of the variance in it's analysis.
Note that the $\Delta_{2,\max}$ and $\Delta_{\infty,\max}$ terms are not affected by $R$ as it's analysis does not use $\rho$.

\section{Proofs for ~\Cref{sec:one_bit}}
\label{sec:one_bit_proofs}

\subsection{Proof of ~\Cref{lem:label flip}}
\label{sec:label_flip_proof}
To prove this Lemma, note that $\tilde{b}_i = sign(\ip{g_i}{z_i}) \neq  sign(\ip{g}{z_i})$ only if  $z_i$ is sampled from the symmetric difference of $g_i$ and $g$. 
The probability that a $z_i$ sampled uniformly from $\mathbb{S}^{d-1}$ lies in this symmteric difference is given by $\nicefrac{\arccos(\ip{g}{g_i})}{\pi}$. If we set $\Delta_{\rm corr} = \frac{1}{m\pi}\sum_{i\in[m]}\arccos(\ip{g}{g_i})$

Let $\zeta$ be the fraction of $z_i$ such that $\tilde{b}_i \neq sign(\ip{g}{z_i})$. Then, by Chernoff bound, we have,
\begin{align*}
    \Pr[\zeta \geq (1 + \gamma) \Delta_{\rm corr} ]\leq e^{- \frac{\gamma^2 m \Delta_{\rm corr}}{2 + \gamma}}
\end{align*}
By setting $\gamma$ to be any small constant, we obtain, with probability $1 - \mathcal{O}(\exp(-m \Delta_{\rm corr}))$, atmost $\zeta = \Theta(\Delta_{\rm corr})$ fraction of datapoints are not generated from the halfspace with normal $g$ and are thus corrupted.

\subsection{Proofs of ~\Cref{lem:one_bit_optimal_bias} and ~\ref{lem:average_bias}}
\label{sec:one_bit_thm_proofs}
To prove ~\Cref{lem:one_bit_optimal_bias}, we utilize the guarantees of  ~\cite[Theorem~3]{pmlr-v206-shen23a}, where the sample complexity requirement ensures that the error is $\tilde{O}(\frac{d}{m})$. Further, ~\cite[Theorem~3]{pmlr-v206-shen23a} obtains error guarantee linear in the noise rate of the samples which is obtained from Lemma~\ref{lem:label flip}. The error guarantee is in terms of the symmetric difference between $\tilde{g}$ and $g$ wrt the uniform distribution on the unit sphere. Since this is equal to the angle between these two vectors divided by $\pi$, this gives us a bound on the inner product of these two unit vectors. 

To prove ~\Cref{lem:average_bias}, from ~\cite[Theorem~12]{kalai_agnostically_2008}, the sample complexity provides the term $\frac{d}{\sqrt{m}}$ while the noise tolerance provides the term $\sqrt{d}\Delta_{\rm corr}$.

\subsection{Proof of ~\Cref{rem:delta_corr}}
\label{sec:delta_corr_proof}
To prove this remark, note that $\arccos(x)$ is concave for $x\geq 0$. Therefore, by applying Jensen's inequality, we obtain,
\begin{align*}
    \Delta_{\rm corr} =& \frac{1}{m\pi}\sum_{i\in [m]}\arccos(\ip{g_i}{g})\leq \frac{1}{\pi}\arccos\left(\ip{\frac{1}{m}\sum_{i=1}^{m}g_i}{g}\right) = \frac{1}{\pi}\arccos\left(\norm{\frac{1}{m}\sum_{i=1}^{m}g_i}_2\ip{g}{g}\right)\\
    \leq& \frac{1}{\pi}\arccos\left(\sqrt{\norm{\frac{1}{m}\sum_{i\in [m]}g_i}_2^2}\right) = \frac{1}{\pi}\arccos\left(\sqrt{\norm{\frac{\sum_{i\in [m]}\ip{g_i}{g_i}}{m^2} + \frac{2}{m^2}\underset{1\le i<j\leq m}{\sum\sum}\ip{g_i}{g_j}}}\right) \\
    =& \frac{1}{\pi}\arccos\left(\sqrt{\frac{1}{m}+\frac{2}{m^2}\underset{1\le i<j\leq m}{\sum\sum}\ip{g_i}{g_j}}\right)
\end{align*}

\section{Additional Experiment Details}
\label{sec:add_exp}
\textbf{Baselines}
We implement all the baselines mentioned in Table~\ref{tab:related_works}. As all these baselines are suited to $\ell_2$ error, for the DME experiment on gaussians, where $\ell_2$ error is the correct metric,  compare SparseReg (Algorithm~\ref{alg:sparc_1}) to all these baselines. For $\ell_\infty$ error uniform distribution, we implement NoisySign (Algorithm~\ref{alg:noisy_sign}) and HadamardMultiDim (Algorithm~\ref{alg:hadamardmultidim}) and compare it to Correlated SRQ~\cite{suresh_correlated_2022}, as it's guarantees hold in single dimensions. We also add comparisons to its independent variant, SRQ~\cite{suresh_distributed_2017}, and Drive~\cite{vargaftik_drive_2021}, which performs coordinate-wise signs. For the unit vector case, we implement OneBit (Algorithm~\ref{alg:onebit} Technique II) and SparseReg(Algorithm~\ref{alg:sparc_1}) and compare it with one independent compressor (SRQ~\cite{suresh_distributed_2017}) and one collaborative compressor (RandKSpatialProj~\cite{jiang_correlation_2023}). Note that we set $d=512$ throughout our experiments and tune the parameters (number of coordinates sent~\cite{konecny_randomized_2018,jhunjhunwala_leveraging_2021} or the quantization levels in ~\cite{suresh_distributed_2017,suresh_correlated_2022}) so that all compressors have the same number of bits communicated. For compressors without tunable parameters, we repeat them to match the communication budget. The communication budget is $2375 \pm 25$ bits/client for each compressor every round. It is often not possible for two different forms of compressors, like sparsification methods and quantization methods to achieve the exact same communication budget even after tuning hence we allow a small range of communication budgets.

\textbf{Datasets} For the distributed mean estimation task, we generate $d$ dimensional vectors on $m=100$ clients. To compare $\ell_2$ error, we generate $g$ with $\norm{g}_2 = 100$. Then, each client generates $g_i$ from a $\cN(0,\Delta_2^2)$, where  $\Delta_2 \in [0.001, 100]$. To compare $\ell_\infty$ error, we generate $g$ uniformly from a hypercube $[-B, B]^d$ where $B=100$. Each client generates $g_i$ from a smaller hypercube $[-\Delta_\infty, \Delta_\infty]^d$ centered at $g$ where $\Delta_\infty \in [10^{-3}, 10^2]$. To compare cosine distance, we generate $g$ uniformly from the unit sphere, and each client generates $g_i$  uniformly from the set of unit vectors at a cosine distance $\Delta_{corr}$ from the $g$, Here, $\Delta_{corr} \in  [0.01, 0.4]$. 

For KMeans and power iteration, we set $m=50$. FEMNIST is a real federated dataset where each client has handwritten digits from a different person. We apply dimensionality reduction to set $d=512$. We run $20$ iterations of Lloyd's algorithm~\cite{lloyds} for KMeans and $30$ power iterations. For distributed linear regression, the Synthetic dataset is a mixture of linear regressions, with one mixture component per client. The true model $w_i\in \R^d$ for each component  is obtained from DME setup for gaussians with $\Delta_2 = 4$.  Then, we generate $n=1000$ datapoints on each client, where the features $x$ are sampled from standard normal, while the labels $y$ are generated as $y = \ip{w_i}{x} + \xi$, where $\xi$ is the zero-mean gaussian noise with variance $10^{-2}$. For UJIndoorLoc, we use the first $d=512$ of the $520$ features following ~\cite{jiang_correlation_2023}. The task for UJIndoorLoc dataset is to predict the longitude of a phone call. For both the linear regression datasets, we run $50$ iterations of GD. For MNIST and UJIndoorLoc, we split the dataset uniformly into $m$ chunks one per client.

For logistic regression for binary classification, we select the last $2$ classes of the HAR and label them with $\pm1$. We split the dataset into $m=20$ clients iid. HAR dataset has $561$ features which we reduce by PCA to $d=512$. We run distributed gradient descent with learning rate $0.001$ for $T=200$ iterations on the logistic loss, where the logistic loss for any data point $(x, y) \in \R^d \times \{\pm 1\}$ is defined as $\ell(w, (x,y))  = \log(1 + \exp( - \ip{w}{x}\cdot y) $ for any weight $w\in \R^d$.

\textbf{Metrics}  With the same number of bits, we can directly compare the error of baselines. For mean estimation, we measure $\ell_2$ error, $\ell_\infty$ error and cosine distance for gaussian, uniform and unit vectors respectively. For KMeans, we report the KMeans objective. For power iteration, we report the top eigenvalue. For linear regression, we provide the mean squared error on a test dataset. For logistic regression, we report the training logistic loss and test accuracy for binary classification. For all experiments except power iteration and test accuracy logistic regression, lower value of the reported metric implies better performance. For power iteration, higher implies better performance, as we need to find the eigenvector corresponding to the top eigenvalue. For logistic regression, higher test accuracy implies better performance.

We provide the code in the supplementary material and all the experiments took $6$ days to run on a single $20$ core machine with $25$ GB RAM.

\section{Distributed Gradient Descent with SparseReg Compressor}
\label{sec:conv_dgd}
This section uses our $\ell_2$ compressor, SparseReg, for running FedAvg. Each client $i\in [m]$ contains a local objective function $f_i:\cW\to \R$. We define the global objective function $f(w) = \frac{1}{m}\sum_{i=1}^m f_i(w), \,\forall w\in \cW \subset \R^d$. The goal is to find $w^\star \in \argmin_{w\in \cW} f(w)$. Note that $\nabla f(w) = \frac{1}{m}\sum_{i=1}^m \nabla f_i(w)$, therefore, in our case, the vector $g_i$ correspond to $\nabla f_i(w)$. We describe the algorithm in Algorithm~\ref{alg:fedavg_sparsereg}

\begin{algorithm}
    \caption{Distributed Projected Gradient Descent with  SparseReg compressor}
    \label{alg:fedavg_sparsereg}
    \begin{algorithmic}
        \REQUIRE Initial iterate $w^0\in \cW$, Step size $\gamma >0$
        \STATE \underline{\texttt{Server}}
        \STATE SparseReg-\texttt{Init()}
        \FOR{$t=0$ to $T-1$}
        \STATE Send $w^t$ to all clients $i\in [m]$.
        \STATE Receive $\tilde{b_i}^t$ from clients $i\in [m]$.
        \STATE $\tilde{g}^t \gets$ SparseReg-\texttt{Decode($\{\tilde{b}_i^t\}_{i\in [m]}$)}
        \STATE $w^{t+1}  \gets \mathrm{proj}_{\cW}(w^t - \eta_t \tilde{g}^t)$
        \ENDFOR 
        \STATE \underline{\texttt{Client(i)} at iteration $t$}
        \STATE Receive $w^t$ from server.
        \STATE $\tilde{b}_i \gets$ SparseReg-\texttt{Encode}($\nabla f_i(w^t)$)
        \STATE Send $\tilde{b}_i^t$ to server.
    \end{algorithmic}
\end{algorithm}
We first state the assumptions required for applying the SparseReg compressor.
\begin{assumption}[Bounded Gradient]
    \label{assump:bdd_grad_fedavg}
    For all $w\in \cW, i\in [m]$, we assume that $\norm{\nabla f_i(w)}_2\leq B$.
\end{assumption}
By this assumption, we ensure that for each iteration $t$ in Algorithm~\ref{alg:fedavg_sparsereg},  $\norm{g_i}_2 = \norm{\nabla f_i(w^t)}_2$ is bounded. Further, bounded gradients imply that each $f_i$ is Lipschitz. By triangle inequality, we can also establish the following corollary.
\begin{corollary}\label{corr:lip}
The objective function $f(w)$ is $B$-Lipschitz, $\forall w\in \cW$.    
\end{corollary}

From the above assumptions, it is clear that local objective functions need to be Lipschitz. From ~\cite[Theorem~3.2]{bubeck_cvx_book}, if the domain of iterates, $\cW$ is bounded and $f(w)$ is also convex, then gradient descent can converge at a rate $\cO(1/\sqrt{T})$. We use these two assumptions, and establish a $\cO(1/\sqrt{T})$ rate along with a error obtained from Theorem~\ref{thm:sparsereg}. We define $\Delta_{\rm reg}(t)$ and $\Delta_{2,\max}(t)$ from Theorem~\ref{thm:sparsereg} to be the corresponding errors for $g_i = \nabla f_i(w^t),\forall i \in [m]$ for any $t>0$.

\begin{assumption}[Bounded domain]\label{assump:bdd_domain}
The set $\cW$ is closed and convex with diameter $R^2$.    
\end{assumption}
\begin{assumption}[Convexity]\label{assump:cvx}
    The objective function $f(w)$ is convex $\forall w\in \cW$.
\end{assumption}

We now state our convergence result.
\begin{theorem}\label{thm:dgd_sparsereg}
    Under Assumptions~\ref{assump:bdd_grad_fedavg}, ~\ref{assump:bdd_domain}, ~\ref{assump:cvx}, running Algorithm~\ref{alg:fedavg_sparsereg} for $T$ iterations with step size $\eta_t = \frac{R}{B\sqrt{T}}$, with probability $1 - 2m^2LT\exp(-d\delta_1^2/8) - mT\left(\frac{L^{2\delta_2}}{\log L}\right)^{-m}$
    we have,
    \begin{equation}
        \begin{aligned}
            &\E[f(\bar{w}^T)] - f(w^\star) \leq \frac{ R(2 B^2 + \Gamma_1)}{2B\sqrt{T}} + \sqrt{\Gamma_1}R ,\quad \text{ where, }\quad  \bar{w}^T = \frac{1}{T} \sum_{t=0}^{T-1}w^t\\
             &\Gamma_1 = B^2\left(1 + \frac{10\log L}{d}\exp\left(\frac{m\log L}{d}\right)(\delta_1 + \delta_2) \right)^2\left(1 - \frac{2\log L}{d}\right)^{m},\\  
            &\Gamma_2 = \max_{t\in \{0,1,\ldots,T-1\}} \,\,\quad \min\{\Delta_{\rm reg}(t), \Delta_{2,\max}(t)\}
        \end{aligned}
    \end{equation}
\end{theorem}
From the above theorem, we can see that the high probability terms and $\Gamma_1$ and $\Gamma_2$ are obtained from Theorem~\ref{thm:sparsereg}. Note that $\Gamma = \cO(B^2 \exp(-m/d))$, therefore, for large $m$, the additional bias term of $R\sqrt{\Gamma_1}$ is very small. Further, the term $\Gamma_2 \leq B^2$, therefore, $\Gamma_2$ only affects constant terms in the convergence rate due to $\sqrt{T}$ in the denominator. If $\exp(-m/d) = \cO(1/\sqrt{T})$ or $m = \Omega(d\log T)$, the final convergence rate of Algorithm~\ref{alg:fedavg_sparsereg} is  $\cO(RB/\sqrt{T})$ which is the rate for distributed GD without compression.

We provide the proof for the above theorem, which modifies the proof of ~\cite[Theorem~3.2]{bubeck_cvx_book} to handle a biased gradient oracle. We can also extend our analysis to other function classes, for instance strongly convex functions, by using existing works on biased gradient oracles~\cite{biased_gradients}. Extending the proof to FedAvg from distributed GD would require using biased gradient oracles in ~\cite{Li2020On}. Further, these proofs can also be extended to HadamardMultiDim compressor, with an additional $\sqrt{d}$ factor in the corresponding error terms from Theorem~\ref{thm:hadamard_error} to account for conversion from $\ell_\infty$ to $\ell_2$ norm.

\subsection{Proof of Theorem~\ref{thm:dgd_sparsereg}}

At any iteration $t>0$, we use $\tilde{g}^{t}$ to denote the estimate of $\nabla f(w^t)$. From the proof of Theorem~\ref{thm:sparsereg}, $\norm{\E_t[\tilde{g}^t] - \nabla f(w^t)}_2 \leq \sqrt{\Gamma_1}, $ and $\mathbb{V}ar_t(\tilde{g}^t|w^t) \leq \Gamma_2, \forall t>0$, where $\E_t$ and $\mathbb{V}ar_t$ are the expectation and variance wrt the randomness in the SparseReg compressor at iteration $t$.  We take a union bound over the high probability terms in Theorem~\ref{thm:sparsereg} over all iterations $t=0$ to $T-1$.

We can write the following equation by convexity of $f(w^t)$.
\begin{align*}
    f(w^t) - f(w^\star) \leq& \ip{\nabla f(w^t)}{w^t - w^\star}= \ip{\tilde{g}^t}{w^t - w^\star} + \ip{\nabla f(w^t) - \tilde{g}^t}{w^t - w^\star} \\
    \leq & \frac{1}{2\eta}(\norm{w^t - w^\star}_2^2 - \norm{w^t - \eta \tilde{g}^t - w^\star}_2^2) + \eta\norm{\tilde{g}^t}_2^2 /2 + \ip{\nabla f(w^t) - \tilde{g}^t}{w^t - w^\star}
\end{align*}
In the second line, we use $2\ip{a}{b} = \norm{a}_2^2 + \norm{b}_2^2 - \norm{a - b}_2^2 $. Now, taking expectation wrt the randomness in SparseReg at iteration $t$, we obtain,
\begin{align*}
    \E_t[f(w^t)] - f(w^\star) \leq & \frac{1}{2\eta}(\norm{w^t - w^\star}_2^2 - \E_t[\norm{w^t - \eta \tilde{g}^t - w^\star}_2^2]) + \eta\E_t[\norm{\tilde{g}^t}_2^2]/2 \\
    & \quad + \ip{\nabla f(w^t) - \E_t[\tilde{g}^t]}{w^t - w^\star}\\
    \leq & \frac{1}{2\eta}(\norm{w^t - w^\star}_2^2 - \E_t[\norm{w^{t+1}- w^\star}_2^2]) + \eta(\norm{\E_t[\tilde{g}^t]}_2^2 + \mathbb{V}ar_t(\tilde{g}^t))/2 \\
    & \quad + \norm{\nabla f(w^t) - \E_t[\tilde{g}^t]}_2\cdot\norm{w^t - w^\star}_2\\
    \leq & \frac{1}{2\eta}(\norm{w^t - w^\star}_2^2 - \E_t[\norm{w^{t+1}- w^\star}_2^2]) + \eta(B^2 + \Gamma_2)/2 + \sqrt{\Gamma_1}R\\
\end{align*}
In the second line, we use the non-expansiveness of projections on a convex set, $\norm{w^t - \eta \tilde{g}^t - w^\star}_2 \geq \norm{\mathrm{proj}_{\cW}(w^t - \eta \tilde{g}^t - w^\star)}_2$, the decomposition of $2^{nd}$ moment into square of mean and variance, and cauchy-schwartz inequality. In the third line, we plug in bounds of $\Gamma_1, \Gamma_2$, diameter of the set and by triangle inequality, argue that $\E[\tilde{g}^t]$ also lies in an $\ell_2$ ball of radius $B$.

Finally, we take expectations wrt all random variables, unroll the recursion from $t=0$ to $T$, and divide both sides by $T$.
\begin{align*}
    \frac{1}{T}\sum_{t=0}^T\E[f(w^t)] - f(w^\star) \leq \frac{R^2}{2\eta T} + \frac{\eta(B^2 + \Gamma_2)}{2} + \sqrt{\Gamma_1}R \leq \frac{ R(2 B^2 + \Gamma_1)}{2B\sqrt{T}} + \sqrt{\Gamma_1}R 
\end{align*}
We obtain the final inequality by plugging in the step size $\eta = \frac{R}{B\sqrt{T}}$. By convexity of $f$, for $\bar{w}^T = \sum_{t=0}^{T-1} w^t$, we obtain,
\begin{align*}
    \E[f(\bar{w}^T)] - f(w^\star) \leq \frac{1}{T}\sum_{t=0}^{T-1} \E[f(w^t)] -f(w^\star) \leq \frac{ R(2 B^2 + \Gamma_1)}{2B\sqrt{T}} + \sqrt{\Gamma_1}R
\end{align*}

\end{document}